\documentclass[a4paper, twoside, 12pt]{article}

\usepackage[noblocks]{authblk} 
\usepackage[numbers]{natbib}
\usepackage{graphicx}
\usepackage{fancyhdr}
\usepackage[colorlinks=true, linkcolor=MidnightBlue, urlcolor=MidnightBlue, citecolor=PineGreen]{hyperref}
\usepackage[tableposition=top, justification=justified, textfont=it]{caption}
\usepackage{titlesec}
\usepackage{multicol}
\usepackage{afterpage}
\usepackage{lipsum}
\usepackage[dvipsnames, x11names]{xcolor}
\usepackage[inner=4cm, outer=4cm, top=2.8cm, bottom=2.8cm, marginparsep=0cm, marginparwidth=0cm, includehead, includefoot]{geometry}
\usepackage{setspace}
\usepackage{array}
\usepackage{bm}
\usepackage{tikz}
\usepackage{float}
\usepackage[framemethod]{mdframed}
\usepackage{amsmath}
\usepackage{amssymb}
\usepackage{comment}

\titleformat{\section}{\Large \bfseries \centering \scshape}{\thesection.}{0.3em}{}[{\titlerule[0.5pt]}]

\definecolor{shadecolor}{RGB}{230,230,230}
\newcommand{\mybox}[1]{\par\noindent\colorbox{shadecolor}
{\parbox{\dimexpr\textwidth-2\fboxsep\relax}{#1}}}

\titleformat{\subsection}{\large \bfseries \mybox}{\thesubsection}{1em}{}

\titleformat{\subsubsection}{\itshape}{\thesubsubsection.}{0.3em}{}

\renewenvironment{abstract}
{\vskip 2.5ex {\large\bf\noindent Abstract}\vspace{0.7ex} \\ %
  \bgroup\noindent\ignorespaces}%
{\par\egroup\vskip 2.5ex}

\newenvironment{keywords}
{\bgroup\leftskip 20pt\rightskip 20pt \small\noindent{\bf Keywords:} }%
{\par\egroup\vskip 10ex}

\fancypagestyle{firstpage}{%
  
  \fancyhead[]{}
  \fancyfoot[C]{}
  \fancyfoot[L]{\footnotesize To appear in \\
  O. Colliot (Ed.), \textit{Machine Learning for Brain Disorders}, Springer\\
  }

}

\fancypagestyle{otherpages}{%
  
  \fancyhead[LE]{\thepage}
  \fancyhead[RE]{\runningauthor}
  \fancyhead[LO]{\runningheadtitle}
  \fancyhead[RO]{\thepage}
  \fancyfoot[L]{\footnotesize  \textit{Machine Learning for Brain Disorders}, Chapter~\chapternumber}
  \fancyfoot[C]{}
}

\makeatletter
\renewcommand{\maketitle}{\bgroup\setlength{\parindent}{0pt}

\begin{flushright}
  \color{MidnightBlue}
  \textbf{\LARGE Chapter~\chapternumber}
\end{flushright}

\vspace{0.3in}

\begin{flushleft}
    \setstretch{2.0} 
    \textbf{\color{MidnightBlue}\huge\@title}
\end{flushleft}

\vspace{0.15in}

\begin{flushleft}
    \textbf{\bfseries \large\@author}
\end{flushleft}\egroup
}

\makeatother

\DeclareCaptionLabelFormat{labelformat}{\color{MidnightBlue}\bfseries #1 #2}
\renewcommand{\bibpreamble}{\scriptsize \begin{multicols}{2}}
\renewcommand{\bibpostamble}{\end{multicols}}

\mdfsetup{skipabove=\topskip, skipbelow=\topskip}

\newcounter{nicebox}

\newfloat{floatbox}{thp}{lop}
\floatname{floatbox}{Box}

\DeclareMathAlphabet{\mathsfit}{\encodingdefault}{\sfdefault}{m}{sl}
\SetMathAlphabet{\mathsfit}{bold}{\encodingdefault}{\sfdefault}{bx}{n}

\begin{document}


\newcommand{\runningauthor}{R. Courant \textit{et al.}} 

\newcommand{\runningheadtitle}{Transformers}

\newcommand{\chapternumber}{6}

\newcommand{\emailaddress}{vicky.kalogeiton@lix.polytechnique.fr}

\title{Transformers and visual Transformers} 

\author[1,2]{Robin Courant}
\author[1]{Maika Edberg}
\author[1]{Nicolas Dufour}
\author[*,1]{Vicky Kalogeiton}  

\affil[1]{LIX, CNRS, Ecole Polytechnique, IP Paris}
\affil[2]{Univ.\ Rennes, CNRS, IRISA, INRIA}

\affil[*]{Corresponding author: e-mail address: \href{mailto:\emailaddress}{\emailaddress}}

\maketitle

\afterpage{\aftergroup\restoregeometry}
\pagestyle{otherpages}

\begin{abstract}
Transformers were initially introduced for natural language processing (NLP) tasks, but fast they were adopted by most deep learning fields, including computer vision. They measure the relationships between pairs of input tokens (words in the case of text strings, parts of images for visual Transformers), termed attention. The cost is exponential with the number of tokens. For image classification, the most common Transformer Architecture uses only the Transformer Encoder in order to transform the various input tokens. However, there are also numerous other applications in which the decoder part of the traditional Transformer Architecture is also used. Here, we first introduce the Attention mechanism (Section~\ref{sec:attention}), and then the Basic Transformer Block including the Vision Transformer (Section~\ref{sec:transformer}). Next, we discuss some improvements of visual Transformers to account for small datasets or less computation (Section~\ref{sec:extensions}). Finally, we introduce Visual Transformers applied to tasks other than image classification, such as detection, segmentation, generation and training without labels (Section~\ref{sec:tasks}) and other domains, such as  video or multimodality using text or audio data (Section~\ref{sec:domains}). 

\end{abstract}


\section{Attention} 
\label{sec:attention}
Attention is a technique in Computer Science that imitates the way in which the brain can focus on the relevant parts of the input. In this section, we introduce attention: its history (Section~\ref{sub:history_attention}), its definition (Section~\ref{sub:definition_attention}), its types and variations  (Sections~\ref{sub:attention_types} and \ref{sub:variation_attention}) and its properties (Section~\ref{sub:property_attention}). \\

To understand what attention is and why it is so useful, consider the following film review:
\begin{center}
	\textit{While others claim the story is boring, I found it fascinating.}
\end{center}
Is this film review positive or negative?   
The first part of the sentence is unrelated to the critic's opinion, while the second part suggests a positive sentiment with the word ‘fascinating’. 
To a human, the answer is obvious; however, this type of analysis is not necessarily obvious to a computer. 

Typically, sequential data require \emph{context} to be understood. 
In natural language, a word has a meaning because of its position in the sentence, with respect to the other words: its \emph{context}. 
In our example, while “boring” alone suggests that the review is negative, its contextual relationship with the other words allows the reader to reach the appropriate conclusion. 
In computer vision, in a task like object detection, the nature of a pixel alone cannot be identified: we need to account for its neighbourhood, its \emph{context}. 
So, how can we formalize the concept of \emph{context} in sequential data?

\subsection{The History of Attention}
\label{sub:history_attention}

This notion of \emph{context} is the motivation behind the introduction of the attention mechanism in 2015 \cite{Bahdanau2015NeuralMT}. 
Before this, language translation was mostly relying on encoder-decoder architectures:  recurrent neural networks (RNNs) ~\cite{cho_RNN_translation_2014} and in particular long short-term memory (LSTMs) networks were used to model the relationship among words~\cite{ Sutskever_LSTM_Translation_2014}. 
Specifically, each word of an input sentence is processed by the encoder sequentially. At each step, the past and present information are summarized and encoded into a fixed-length vector. 
In the end, the encoder has processed every word, and outputs a final fixed-length vector, which summarizes all input information. This final vector is then decoded, and finally translates the input information into the target language.

However, the main issue of such structure is that all the information is compressed into one fixed-length vector. 
Given that the sizes of sentences vary and as the sentences get longer, a fixed-length vector is a real bottleneck: it gets increasingly difficult not to lose any information in the encoding process due to the vanishing gradient problem~\cite{Bahdanau2015NeuralMT}. 

As a solution to this issue, Bahdanue et al.~\cite{Bahdanau2015NeuralMT} proposed the attention module in 2015. 
The attention module allows the model to consider the parts of the sentence that are relevant to predicting the next word. Moreover, this facilitates the understanding of relationships among words that are further apart.

\subsection{Definition of Attention}
\label{sub:definition_attention}
\begin{figure}
    \centering
    \includegraphics[scale=0.8]{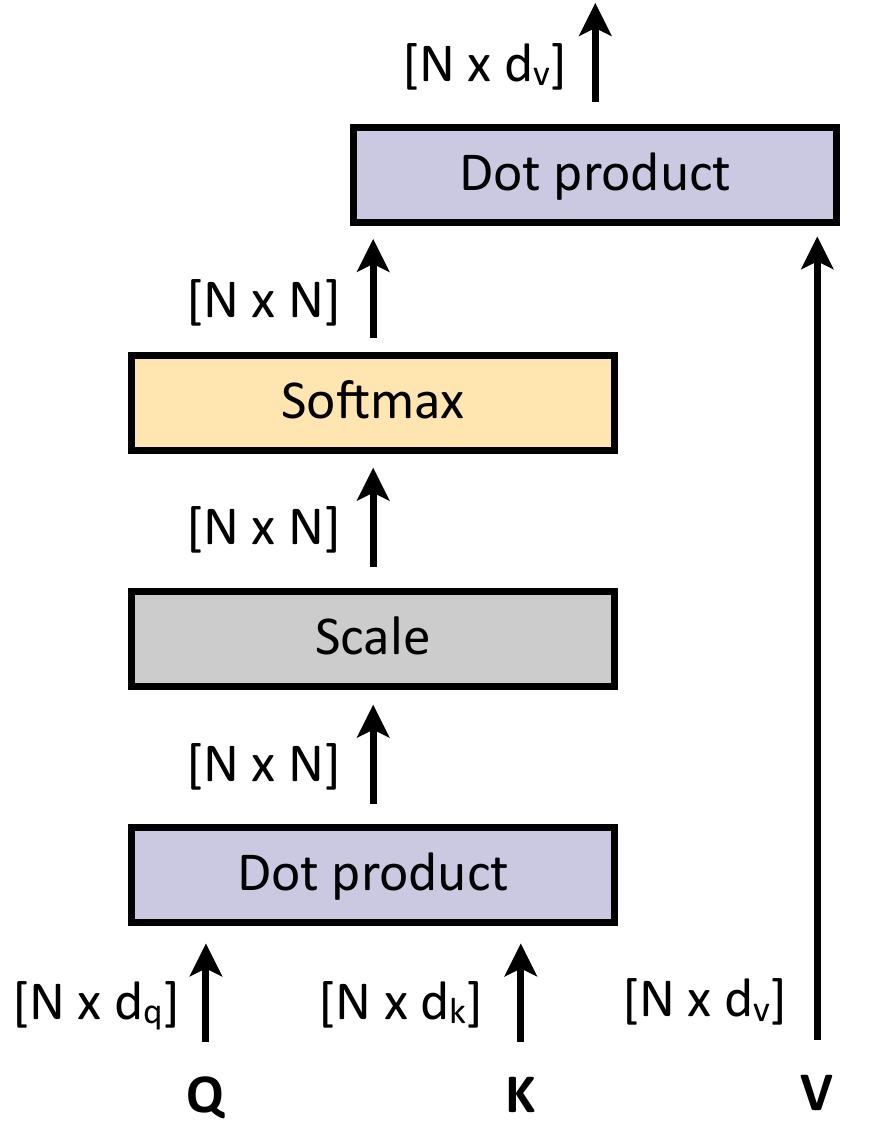}
    \caption{\textbf{Attention block.} Next to each element, we denote its dimensionality. Figure inspired from~\cite{vaswani2017attention}.  }
    \label{fig:attention}
\end{figure}

Given two lists of tokens, $\bm{X}\in\mathbb{R}^{N \times d_x}$ and $\bm{Y}\in\mathbb{R}^{N \times d_y}$, attention encodes information from $\bm{Y}$ into $\bm{X}$, where $N$ is the length of inputs $\bm{X}$ and $\bm{Y}$, and $d_x$ and $d_y$ are their respective dimensions. 
For this, we first define three linear mappings: 
query mapping $\bm{W}^Q\in\mathbb{R}^{d_x \times d_{q}}$,  key mapping $\bm{W}^K\in\mathbb{R}^{d_y \times d_{k}}$ and value mapping $\bm{W}^V\in\mathbb{R}^{d_y \times d_{v}}$, where $d_q$, $d_k$, and $d_v$ is the embedding dimension in which the query, key, and value are going to be computed, respectively. 

Then, we define the query $\bm{Q}$, key $\bm{K}$ and value $\bm{V}$~\cite{vaswani2017attention} as:
\begin{align*}
    \bm{Q} &= \bm{X} \bm{W}^Q \\
    \bm{K} &= \bm{Y} \bm{W}^K \\
    \bm{V} &= \bm{Y} \bm{W}^V 
\end{align*}

Next, the \emph{attention matrix} is defined as:
\begin{equation}
    \label{eq:attention_matrix}
    A(\bm{Q}, \bm{K}) = \text{Softmax}
    \left(
    \frac{\bm{Q}\bm{K}^\top}{\sqrt{d_k}}
    \right) \quad . 
\end{equation}

This is illustrated in the left part of Figure~\ref{fig:attention}. The nominator $\bm{Q K}^T \in \mathbb{R}^{N \times N}$ represents how each part of the input in $\bm{X}$ attends to each part of the input in $\bm{Y}$\footnote{Note that in the literature, there are two main attention functions: additive attention~\cite{Bahdanau2015NeuralMT} and dot-product attention (Equation~\ref{eq:attention_matrix}). In practice, the dot-product is more efficient since it is implemented using highly optimized matrix multiplication, compared to the feed forward network of the additive attention; hence, the dot-product is the dominant one.}. This dot-product is then put through the Softmax function to normalize its values and get positive values that add to 1. However, for large values of $d_k$, this may result in the Softmax to have incredibly small gradients, so it is scaled down by $\sqrt{d_k}$. 

The resulting $N \times N$ matrix encodes the relationship between $\bm{X}$ with respect to $\bm{Y}$: it measures how important a token in $\bm{X}$ is with respect to another one in $\bm{Y}$.

Finally, the \emph{attention output} is defined as:
\begin{equation}
    \label{eq:attention_output}
    \text{Attention}(\bm{Q},\bm{K}, \bm{V})= A(\bm{Q}, \bm{K}) \bm{V} \quad .
\end{equation}

Figure~\ref{fig:attention} displays this.  
The attention output encodes the information of each token by taking into account the contextual information. 
Therefore, through the learnable parameters -- queries, keys, and values--, the attention layers learn a token embedding that takes into account their relationship. 

\paragraph{Contextual relationships.}
How does Equation~\ref{eq:attention_output} encode contextual relationships? To answer this question, let us reconsider analysing the sentiment of film reviews. To encode contextual relationships into the word embedding, we first want a matrix representation of the relationship between all words. To do so, given a sentence of length $N$, we take each word vector and feed it to two different linear layers, calling one output `query' and the other output `key'. We pack the queries into the matrix $\bm{Q}$ and the keys into the matrix $\bm{K}$, by taking their product ($\bm{Q} \bm{K}^T$). The result is a $N \times N$ matrix that explains how important the $i$-th word (row-wise) is to understand the $j$-th word (column-wise). This matrix is then scaled and normalized by the division and Softmax. Next, we feed the word vectors into another linear layer, calling its output `value'.   
We multiply these two matrices together. The result of their product are attention vectors that encode the meaning of each word, by including their contextual meaning as well. Given that each of these queries, keys, and values are learnable parameters, as the attention layer is trained, the model learns how relationships among words are encoded in the data. 

\subsection{Types of Attention}
\label{sub:attention_types}

There exist two dominant types of attention mechanisms: \emph{self-attention} and  \emph{cross-attention} \cite{vaswani2017attention}. 
In \emph{self-attention}, the queries, keys and values come from the same input, i.e., $\bm{X} = \bm{Y}$; 
in \emph{cross-attention}, the queries come from a different input than the key and value vectors, i.e., $\bm{X} \neq \bm{Y}$. 
These are described below in Sections~\ref{subsub:self_attention} and \ref{subsub:cross_attention}, respectively.  

\subsubsection{Self-Attention}
\label{subsub:self_attention}

In self-attention, the tokens of $\bm{X}$ attend to themselves ($\bm{X}=\bm{Y}$). Therefore, it is modelled as follows: 
\begin{equation}
    \text{SA}(\bm{X}) = \text{Attention}(
                           \bm{X}\bm{W}^Q , 
                           \bm{X}\bm{W}^K, 
                           \bm{X}\bm{W}^V ) \quad .
\end{equation}

Self-attention formalizes the concept of context. It learns the patterns underlying how parts of the input correspond to each other. By gathering information from the same set, given a sequence of tokens, a token can attend to its neighbouring tokens to compute its output.

\subsubsection{Cross-Attention}
\label{subsub:cross_attention}

Most real-world data are multimodal -- for instance, videos contain frames, audios and subtitles, images come with captions, etc. Therefore, models that can deal with such types of multimodal information have become essential. 

Cross-attention is an attention mechanism designed to handle multimodal inputs.
Unlike self-attention, it extracts queries from one input source and key-value pairs from another one ($\bm{X} \neq \bm{Y}$). It answers the question: `which parts of input $\bm{X}$ and input $\bm{Y}$ correspond to each other?'
Cross-attention (CA) is defined as:
\begin{equation}
    \text{CA}(\bm{X}, \bm{Y}) = \text{Attention}(
        \bm{X}\bm{W}^Q , 
        \bm{Y}\bm{W}^K, 
        \bm{Y}\bm{W}^V) \quad .
\end{equation}

\subsection{Variation of Attention}
\label{sub:variation_attention}

Attention is typically employed in two ways: 
(1) Multi-head Self-Attention (MSA, Section~\ref{subsub:multihead}) and (2) Masked Multi-head Attention (MMA, Section~\ref{subsub:maskedmultihead}). 

\paragraph{Attention Head.}
We call Attention Head the mechanism presented in Section~\ref{sub:definition_attention}, i.e., query-key-value projection, followed by scaled dot product attention (Equations~\ref{eq:attention_matrix} and \ref{eq:attention_output}).

When employing an attention-based model, relying only on a single attention head can inhibit learning. 
Therefore, the Multi-head Attention block is introduced \cite{vaswani2017attention}. 

\begin{figure}
    \centering
    \includegraphics[scale=0.8]{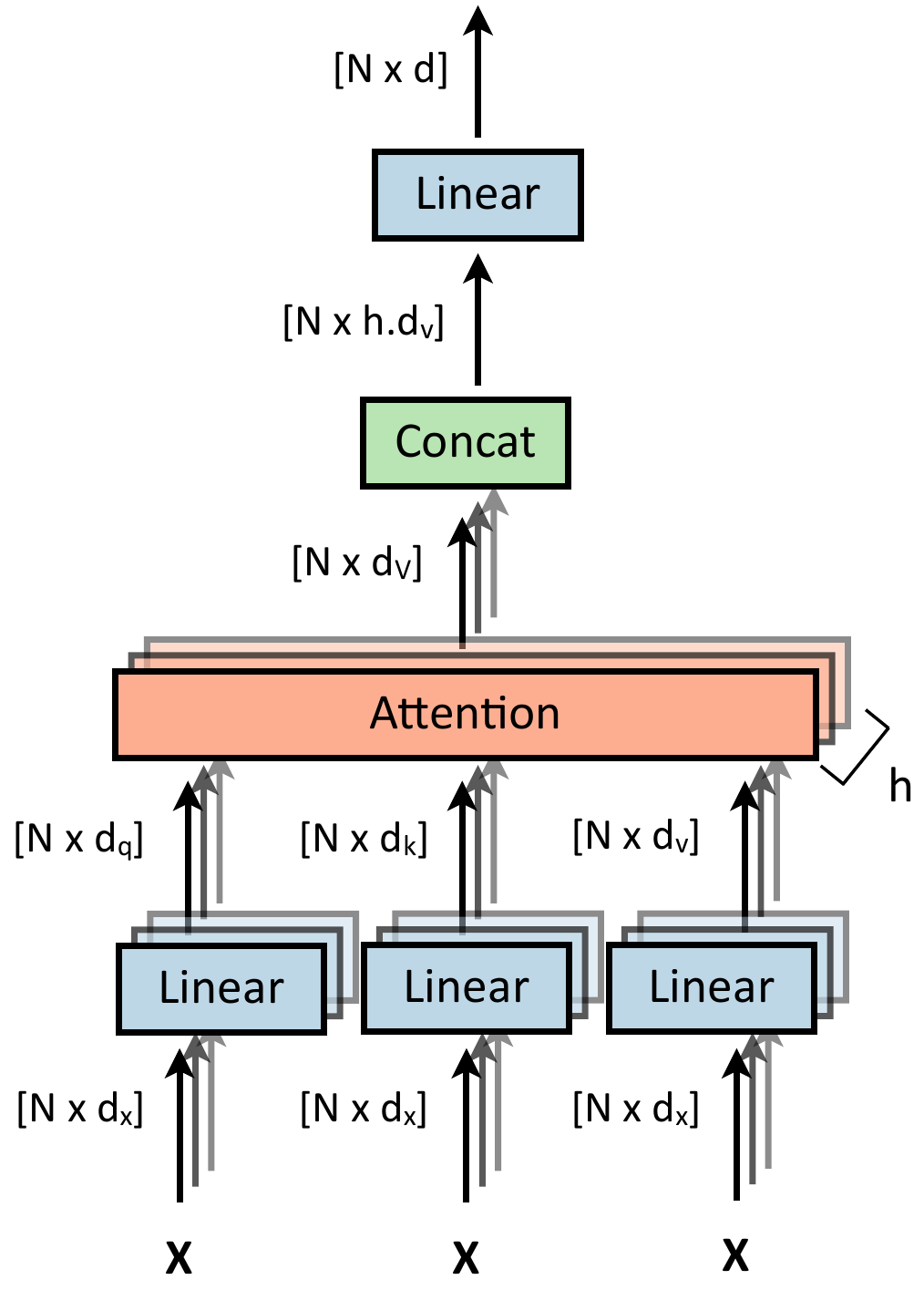}
    \caption{\textbf{Multi-head Self-Attention block (MSA).} 
    First, the input $\bm{X}$ is projected to queries, keys and values and then passed through $h$ attention blocks. The $h$ resulting attention outputs are then concatenated together and finally projected to a $d$-dimensional output vector.
    Next to each element, we denote its dimensionality. Figure inspired from \cite{vaswani2017attention}. 
    }
    \label{fig:multi_head_attention}
\end{figure}

\paragraph{Multi-head Self-Attention (MSA).} 
\label{subsub:multihead}
MSA is shown in Figure~\ref{fig:multi_head_attention} and is defined as: 

\begin{equation}
\label{eq:multihead_self_attention}
\begin{array}{c}
    \text{MSA}(\bm{X}) = \text{Concat} \big( \text{head}_{1}(\bm{X}), \dots, \text{head}_{h}(\bm{X}) \big) \bm{W}^O \quad, \\
     \quad 
    \text{head}_{i}(\bm{X}) = \text{SA}(\bm{X}) \text{ , }\forall i \in \{1,h\} \quad ,
\end{array}
\end{equation}

\noindent where Concat is the concatenation of $h$ attention heads and $\bm{W}^O\in\mathbb{R}^{hd_v \times d}$ is projection matrix. 
This means that the initial embedding dimension $d_x$ is decomposed into $h \times d_v$, and the computation per head is carried out independently. The independent attention heads are usually concatenated and multiplied by a linear layer to match the desired output dimension. The output dimension is often the same as the input embedding dimension $d$. This allows an easier stacking of multiple blocks. 

\paragraph{Multi-head Cross-Attention (MCA).}
Similar to MSA, MCA is defined as:

\begin{equation}
\label{eq:multihead_cross_attention}
\begin{array}{c}
\text{MCA}(\bm{X}, \bm{Y}) = \text{Concat}(\text{head}_1(\bm{X}, \bm{Y}), \dots, \text{head}_h(\bm{X}, \bm{Y}))\bm{W}^O \quad, \\ 
     \quad 
    \text{head}_{i}(\bm{X}, \bm{Y}) = \text{CA}(\bm{X}, \bm{Y}) \text{ , }\forall i \in \{1,h\} \quad .
\end{array}
\end{equation}

\paragraph{Masked Multi-head Self-Attention (MMSA).}
\label{subsub:maskedmultihead}
The MMSA layer~\cite{vaswani2017attention} is another variation of attention. It has the same structure as the Multi-head Self-Attention block (Section~\ref{subsub:multihead}), but all the later vectors in the target output are masked. When dealing with sequential data, this can help make training parallel. 

\subsection{Properties of Attention}
\label{sub:property_attention}

While attention encodes contextual relationships, it is \emph{permutation equivalent}, as the mechanism does not account for the order of the input data. As shown in Equation~\ref{eq:attention_output}, the attention computations are all matrix multiplication and normalizations. Therefore, a permuted input results in a permuted output. In practice, however, this may not be an accurate representation of the information. For instance, consider the sentences `the monkey ate the banana' and `the banana ate the monkey'. They have distinct meanings because of the order of the words. 
If the order of the input is important, various mechanisms, such as the Positional Encoding, discussed in Section~\ref{subsub:positional_encoding}, are used to capture this subtlety. 

\section{Visual Transformers} 
\label{sec:transformer}
The Transformer architecture was introduced in~\cite{vaswani2017attention} and is the first architecture that relies purely on attention to draw connections between the inputs and outputs. 
Since its debut, it revolutionized Deep Learning, making breakthroughs in numerous fields, including Natural Language Processing, Computer Vision, Chemistry, Biology, thus making its way to becoming the \textit{default} architecture for learning representations. 
Recently, the standard Transformer~\cite{vaswani2017attention} has been adapted for vision tasks~\cite{dosovitskiy2020vit}. And again, visual Transformer has become one of the central architectures in computer vision.

In this section, we first introduce the basic architecture of Transformers (Section~\ref{sub:basic_transformer}) and then present its advantages (Section~\ref{sub:advantage_transformers}). Finally, we describe the Vision Transformer (Section~\ref{sub:vit}).

\subsection{Basic Transformers}
\label{sub:basic_transformer}

\begin{figure}[hbtp]
    \centering
    \includegraphics[width=0.8\textwidth]{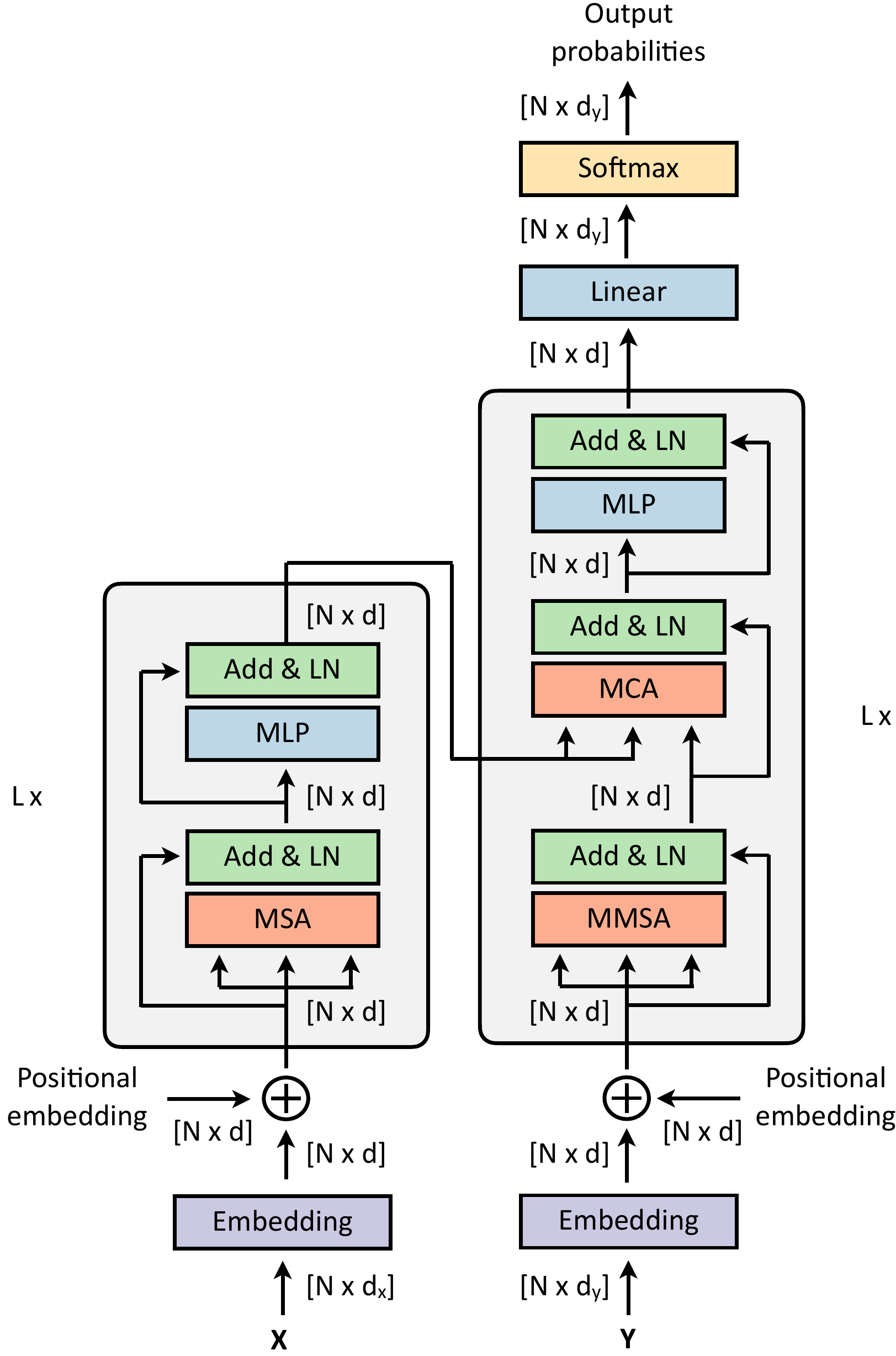}
    \caption{\textbf{The Transformer architecture.} It consists of an encoder (left) and a decoder (right) block, each one consisting from a series of attention blocks (Multi-head and Masked Multi-head attention) and MLP layers. Next to each element, we denote its dimentionality. Figure inspired from~\cite{vaswani2017attention}. 
    }
    \label{fig:transformer_arch}
\end{figure}

As shown in Figure~\ref{fig:transformer_arch}, the Transformer architecture~\cite{vaswani2017attention} is an encoder-decoder model. First, it embeds input tokens $\bm{X} = (\bm{x_1}, \dots, \bm{x_N})$ into a latent space, resulting in latent vectors $\bm{Z} = (\bm{z_1}, \dots, \bm{z_N})$, which are fed to the decoder to output $\bm{Y} = (\bm{y_1}, \dots, \bm{y_M})$. 
The encoder is a stack of $L$ layers, with each one consisting of two sub-blocks: Multi-head Self-Attention (MSA) layers and a Multi-Layer Perceptron (MLP). 
The decoder is also a stack of $L$ layers, with each one consisting of three sub-blocks: Masked Multi-head Self-Attention (MMSA),  Multi-head Cross-Attention (MCA), and a MLP.

\paragraph{Overview.}
Below, we describe the various parts of the Transformer architecture, following Figure~\ref{fig:transformer_arch}. 
First, the input tokens are converted into the embedding tokens (Section~\ref{subsub:embedding}). 
Then, the Positional encoding adds a positional token to each embedding token to denote the order of tokens (Section~\ref{subsub:positional_encoding}). 
Then, the Transformer encoder follows (Section~\ref{subsub:transformer_encoder}). This consists of a stack of $L$ multi-head attention, nomalization and MLP layers and encodes the input to a set of semantically meaningful features. 
After, the decoder follows (Section~\ref{subsub:transformer_decoder}). This consists of a stack of $L$ masked multi-head attention, multi-head attention, and MLP layers followed by normalizations and decodes the input features with respect to the output embedding tokens. 
Finally, the output is projected to linear and Softmax layers. 

\subsubsection{Embedding}
\label{subsub:embedding}

The first step of Transformers consists in converting input tokens\footnote{Note, the initial Transformer architecture was proposed for natural language processing (NLP), and therefore the inputs were words.} into embedding tokens, i.e., vectors with meaningful features. 
To do so, following standard  practice~\cite{word_embedding_Press_2016}, each input is projected into an embedding space to obtain embedding tokens $\bm{Z^{\text{e}}}$. 
The embedding space is structured in a way that the distance between a pair of vectors is relative to the semantic similarity of their associated words. For the initial NLP case, this means that we get a vector of each word, such that the vectors that are closer together have similar meanings.

\subsubsection{Positional Encoding}
\label{subsub:positional_encoding}

As discussed in Section~\ref{sub:property_attention}, the attention mechanism is positional agnostic, which means that it does not store the information on the position of each input. 
However, in most cases, the order of input tokens is relevant and should be taken into account, such as the order of words in a sentence matter as they may change its meaning. 
Therefore, \cite{vaswani2017attention} introduced the \emph{\textbf{P}ositional \textbf{E}ncoding} $\textbf{PE}\in\mathbb{R}^{N \times d_x}$, which adds a positional token to each embedding token $\bm{Z^{\text{e}}} \in\mathbb{R}^{N \times d_x}$. 

\paragraph{Sinusoidal Positional Encoding.}
The Sinusoidal Positional Encoding~\cite{vaswani2017attention} is the main positional encoding method, which encodes the position of each token with sinusoidal waves of multiples frequency. For an embedding token $\bm{Z^{\text{e}}} \in\mathbb{R}^{N\times d_x}$, its Positional Encoding $\textbf{PE}\in\mathbb{R}^{N\times d_x}$ is defined as: 

\begin{equation}
\label{eq:positional_encoding}
\begin{array}{l}
    \textbf{PE}(i,2j)   = \text{sin} \left(\frac{i}{10000^{2j/d}} \right) \\
    \textbf{PE}(i,2j+1) = \text{cos} \left(\frac{i}{10000^{2j/d}} \right), \forall i,j \in [|1,n|]\times[|1,d|] \quad .
\end{array}
\end{equation}

\paragraph{Learnable Positional Encoding.}
An orthogonal approach is to let the model learn the positional encoding. In this case, $\textbf{PE}\in\mathbb{R}^{N\times d_x}$ becomes a learnable parameter. This, however, increases the memory requirements, without necessarily bringing improvements over the sinusoidal encoding.

\paragraph{Positional Embedding.}
After its computation, the Positional Encoding \textbf{PE} is either added to the embedding tokens, or they are concatenated as follows: 

\begin{equation}
\label{eq:positional_embedding}
\begin{array}{l}
    \bm{Z}^{\text{pe}} = \bm{Z^{e}}+\bm{\textbf{PE}} \quad, \text{ or} \\
    \bm{Z}^{\text{pe}} = \text{Concat}(\bm{Z^{e}}, \bm{\textbf{PE}}) \quad, 
\end{array}
\end{equation}

\noindent where Concat denotes vector concatenation. 
Note that the concatenation has the advantage of not altering the information contained in $\bm{Z}^e$, since the positional information is only added to the unused dimension. Nevertheless, it augments the input dimension, leading to higher memory requirements.
Instead, the addition does preserve the same input dimension, while altering the content of the embedding tokens. When the input dimension is high, this content altering is trivial, as most of the content is preserved. 
Therefore, in practice, for high-dimension summing positional encodings is preferred, whereas for low dimensions, concatenating them prevails. 

\subsubsection{Encoder Block}
\label{subsub:transformer_encoder}

The encoder block takes as input the embedding and positional tokens and outputs features of the input, to be decoded by the decoder block. It consists of a stack of $L$ Multi-head Self-Attention (MSA) layers and a Multi-Layer Perceptron (MLP). Specifically, the embedding and positional tokens, $\bm{Z}_{x}^{\text{pe}} \in \mathbb{R}^{N \times d}$, go through a multi-head self-attention block. Then, a residual connection with layer normalization is deployed. In the Transformer, this operation is performed after each sub-layer. Next, we feed its output it to a MLP and a normalization layer. This operation is performed $L$ times and each time the output of each encoder block (of size $N \times d$) is the input of the subsequent block. The $L-$th time, the output of the normalization is the input of the cross attention block in the decoder (Section~\ref{subsub:transformer_decoder}). 

\subsubsection{Decoder Block}
\label{subsub:transformer_decoder}

The decoder has two inputs: first, an input that constitutes the queries $\bm{Q} \in \mathbb{R}^{N \times d}$ of the encoder, and, second, the output of the encoder that constitutes the key-value $\bm{K}, \bm{V} \in \mathbb{R}^{N \times d}$ pair. Similar to Sections~\ref{subsub:embedding}-\ref{subsub:positional_encoding}, the first step constitutes encoding the output token to output embedding token and output positional token. These tokens are fed into the main part of the decoder, which consists of a stack of $L$ Masked Multi-head Self-Attention (MMSA) layers, Multi-Head Cross-Attention (MCA) layers, and Multi-Layer Perceptron (MLP) followed by normalizations. Specifically, the embedding and positional tokens, $\bm{Z}_{y}^{\text{pe}} \in \mathbb{R}^{N \times d}$, go through a MMSA block. Then, a residual connection with layer normalization follow. Next, a MCA layer (followed by a normalization) maps the queries to the encoded keys-values before forwarding the output to a MLP. Finally, we project the output of the $L$ decoder blocks (of dimension $N \times d_y$) through a linear layer and get output probability through a soft-max layer.

\subsection{Advantages of Transformers}
\label{sub:advantage_transformers}

Since their introduction, the Transformers have had a significant impact on deep learning approaches. 

In natural language processing (NLP), before Transformers, most  architectures used to rely on recurrent modules, such as RNNs~\cite{cho_RNN_translation_2014} and in particular LSTMs~\cite{Sutskever_LSTM_Translation_2014}. 
However, recurrent models process the input sequentially, meaning that, to compute the current state, they require the output of the previous state. This makes them tremendously inefficient, as they are impossible to parallelize.
On the contrary, in Transformers, each input is processed independent of the others, and the multi-head attention can perform multiple attention computations at once. This makes Transformers highly efficient, as they are highly parallelizable.  

This results in not only exceptional scalability, both in the complexity of the model and size of datasets, but also relatively fast training. Notably, the recent Switch Transformers~\cite{fedus_switch_2021} was pre-trained on 34 billion tokens from the C4 dataset~\cite{RAffel_C4_dataset_2020}, scaling the model to over 1 trillion parameters.

This scalability~\cite{fedus_switch_2021} is the principal reason for the power of the Transformer. While it was originally introduced for translation, it refrains from introducing many inductive biases, i.e., the set of assumptions that the user makes about the structure of the model input. In doing so, the Transformer relies on data to learn how they are structured. Compared to its counterparts with more biases, the Transformer requires much more data to produce comparable results~\cite{dosovitskiy2020vit}. However, if a sufficient amount of data is available, the lack of inductive bias becomes a strength. By learning the structure of the data from the data, the Transformer is able to learn better without human assumptions hindering~\cite{khan_vision_survey_2021}.

In most tasks involving Transformers, the model is first pre-trained on a large dataset, and then fine-tuned for the task at hand on a smaller dataset. The pre-training phase is essential for Transformers to learn the global structure of the specific input modality. For fine-tuning, typically fewer data suffice as the model is already rich. 
For instance, in natural language processing, BERT~\cite{devlin2018bert}, a state-of-the-art language model, is pre-trained on a Wikipedia-based dataset \cite{wikidump}, with over 6 million articles and Book Corpus \cite{zhu_corpus_2015} with over 10,000 books. Then, this model can be fine-tuned on much more specific tasks.
In Computer Vision, the Vision Transformer (ViT) is pre-trained on the JFT-300M dataset, containing over one billion labels for 300 million images \cite{dosovitskiy2020vit}.
Hence, with a sufficient amount of data, Transformers achieve results that were never possible before in various areas of Machine Learning.

\subsection{Vision Transformer}
\label{sub:vit}

Transformers offer an alternative to CNNs that have long held a stranglehold on Computer Vision.  
Before 2020, most attempts to use Transformers for vision tasks were still highly reliant on CNNs, either by using self-attention jointly with convolutions~\cite{wang_vision_2017,carion_vision_2020} 
or by keeping the general structure of CNNs while using self-attention~\cite{Ramachandran_vision_2019,wang_vision_2020}.

The reason for this is rooted in the two main weaknesses of the Transformers. First, the complexity of the attention operation is high. As attention is a quadratic operation, the number of parameters skyrockets quickly when dealing with visual data, i.e., images --and even more so with videos--. For instance, in the case of ImageNet~\cite{deng2009imagenet}, inputting a single image with $256 \times 256 = 65,536$ pixels in an attention layer would be too heavy computationally. Second, Transformers suffer from lack of inductive biases. Since CNNs were specifically created for vision tasks, their architecture includes spatial inductive biases, like translation equivariance and locality. Therefore, the Transformers have to be pre-trained on a significantly large dataset to achieve similar performances.

\begin{figure}
    \centering
    \includegraphics[width=1.0\textwidth]{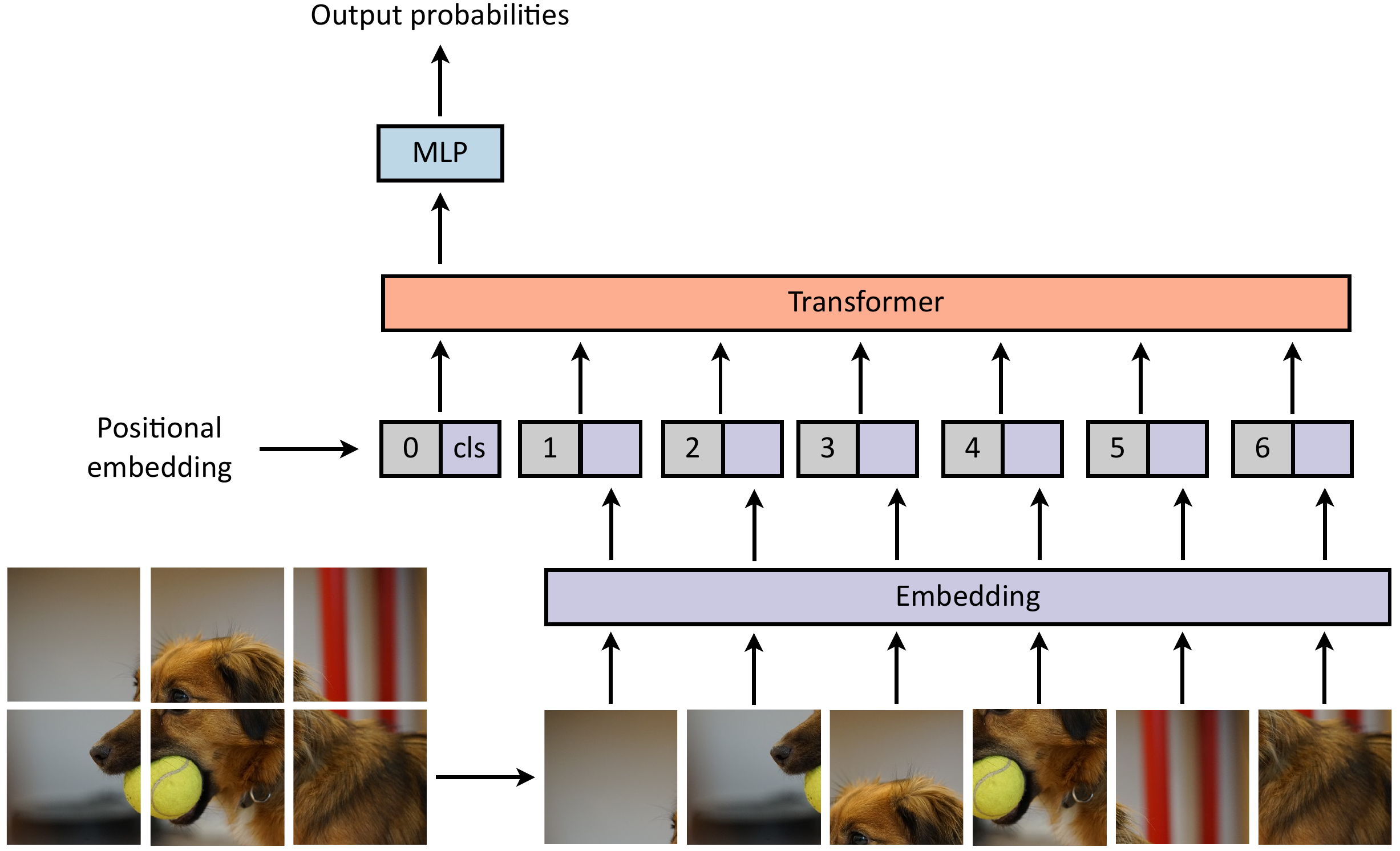}
    \caption{\textbf{The Vision Transformer architecture (ViT).} First, the input image is split into patches (bottom), which are linearly projected (embedding), and then concatenated with positional embedding tokens. The resulting tokens are fed into a Transformer, and finally the resulting classification token is passed through an MLP to compute output probabilities.
    Figure inspired from~\cite{dosovitskiy2020vit}.}
    \label{fig:vit_arch}
\end{figure}

The Vision Transformer (ViT)~\cite{dosovitskiy2020vit} is the first systematic approach that uses directly Transformers for vision tasks by addressing both aforementioned issues. It rids the concept of convolutions altogether, using purely a Transformer-based architecture. In doing so, it achieves the state of the art on image recognition on various datasets, including ImageNet~\cite{deng2009imagenet} and CIFAR-100~\cite{krizhevsky2009cifar}. 

Figure~\ref{fig:vit_arch} illustrates the ViT architecture. 
The input image is first split into $16 \times 16$ patches, flattened and mapped to the expected dimension through a learnable linear projection. Since the image size is reduced to $16 \times 16$, the complexity of the attention mechanism is no longer a bottleneck. Then, ViT encodes the positional information and attaches a learnable embedding to the front of the sequence, similarly to BERT’s classification token~\cite{devlin2018bert}. The output of this token represents the entirety of the input -- it encodes the information from each part of the input. Then, this sequence is fed into an encoder block, with the same structure as in the standard Transformers~\cite{vaswani2017attention}. The output of the classification token is then fed into an MLP that outputs class probabilities.

Due to lack of inductive biases, when ViT is trained only on mid-sized datasets such as ImageNet, it scores some percentage points lower than the state of the art. Therefore, the proposed model is first pre-trained on the JFT-300M dataset~\cite{sun_jft300m_2017} and then fine-tuned on smaller datasets, thereby increasing its accuracy by $13\%$. 

For a complete overview of visual Transformers and follow-up works, we invite the readers to study~\cite{khan_vision_survey_2021,selva_video_survey_2022}.

\section{Improvements over the Vision Transformer} 
\label{sec:extensions}
In this section, we present Transformer-based methods that improve over the original Vision Transformer (Section~\ref{sub:vit}) in two main ways. First, we introduce approaches that are trained on smaller datasets, unlike ViT~\cite{dosovitskiy2020vit} that requires pre-training on 300 million labelled images (Section~\ref{sub:data_efficiency}). 
Second, we present extensions over ViT that are more computational efficient than ViT, given that training a ViT is directly correlated to the image resolution and the number of patches (Section~\ref{sub:compute_efficiency}).

\subsection{Data Efficiency}
\label{sub:data_efficiency}

As discussed in Section~\ref{sub:vit}, the Vision Transformer (ViT)~\cite{dosovitskiy2020vit} is pretrained on a massive proprietary dataset (JFT-300M) which contains 300 million labelled images. This need arises with Transformers because we remove the inductive biases from the architecture compared to convolutional based networks. Indeed, convolutions contain some translation equivariance. ViT does not benefit from this property and thus has to learn such biases, requiring more data. JFT-300M is an enormous dataset and to make ViT work in practice, better data-efficiency is needed. Indeed, collecting that amount of data is costly and can be infeasible for most tasks.

\begin{figure}[hbtp]
	\centering
		\includegraphics[width=0.8\textwidth]{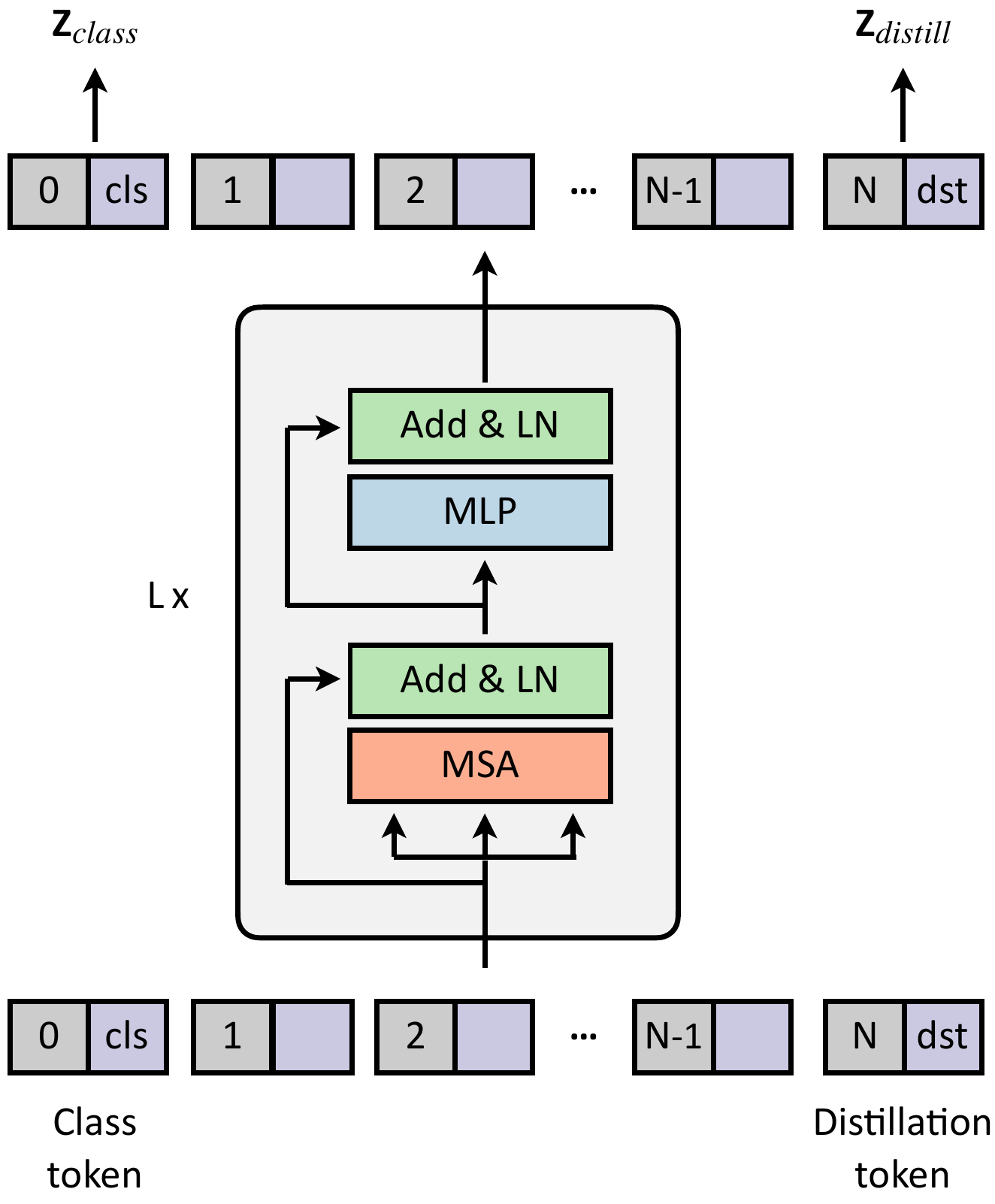}
	   \caption{\textbf{The DeiT architecture.} The architecture features an extra token, the distillation token. This token is used similarly to the class token. Figure inspired from~\cite{deit}.}
	\label{fig:deit}
\end{figure}

\paragraph{Data-efficient image Transformers (DeiT)~\cite{deit}.}
The first work to achieve an improved data efficiency is DeiT~\cite{deit} . The main idea of DeiT is to distil the inductive biases from a CNN into a Transformer (Figure~\ref{fig:deit}). DeiT adds another token that works similarly to the class token. When training, ground truth labels are used to train the network according to the class token output with a cross entropy loss. However, for the distillation network, the output labels are compared to the labels provided from a teacher network with a cross entropy loss. The final loss for a $N$-categorical classification task is defined as follows:
\begin{equation}
\begin{array}{c}
    \mathcal{L}_{global}^{hardDistill} = \frac{1}{2}(\mathcal{L}_{CE}(\Psi(\bm{Z}_{class}), \bm{y}) + \mathcal{L}_{CE}(\Psi(\bm{Z}_{distill}), \bm{y}_T) ) \quad , \\
    \mathcal{L}_{CE}(\bm{\hat{y}}, \bm{y}) = - \frac{1}{N}\sum_{i = 1}^{N} \left[ y_i \log \hat{y}_i + (1 - y_i) \log(1 - \hat{y}_i) \right] 
\end{array}
\end{equation}
with $\Psi$ the Softmax function, $\bm{Z}_{class}$ the class token output,  $\bm{Z}_{distill}$ the class token output, $\bm{y}$ the ground truth label and $\bm{y}_T$ the teacher label prediction. 

The teacher network is a Convolutional Neural Network (CNN). The main idea is that the distillation head will provide the inductive biases needed to improve the data efficiency of the architecture. By doing this, DeiT achieves remarkable performance on the ImageNet dataset, by training `only' on ImageNet-1K~\cite{deng2009imagenet}, which contains 1.3 million images.

\paragraph{Convit~\cite{convit}.}
The main disadvantage of DeiT~\cite{deit} is that is requires a pretrained CNN, which is not ideal, and it would be more convenient to not have this requirement. The CNN has a hard inductive bias constraint that can be a major limitation. Indeed, if enough data is available, learning the biases from the data can result in better representations. 

Convit~\cite{convit} overpasses this issue by including the inductive bias of CNNs into a Transformer in a soft way. Specifically, if the inductive bias is limiting the  training, the Transformer can discard it. The main idea is to include the inductive bias into the ViT initialization. Therefore, before beginning training, the ViT is equivalent to a CNN. Then, the network can progressively learn the needed biases and diverge from the CNN initialization. 

\begin{figure}[hbtp]
	\centering
		\includegraphics[width=\textwidth]{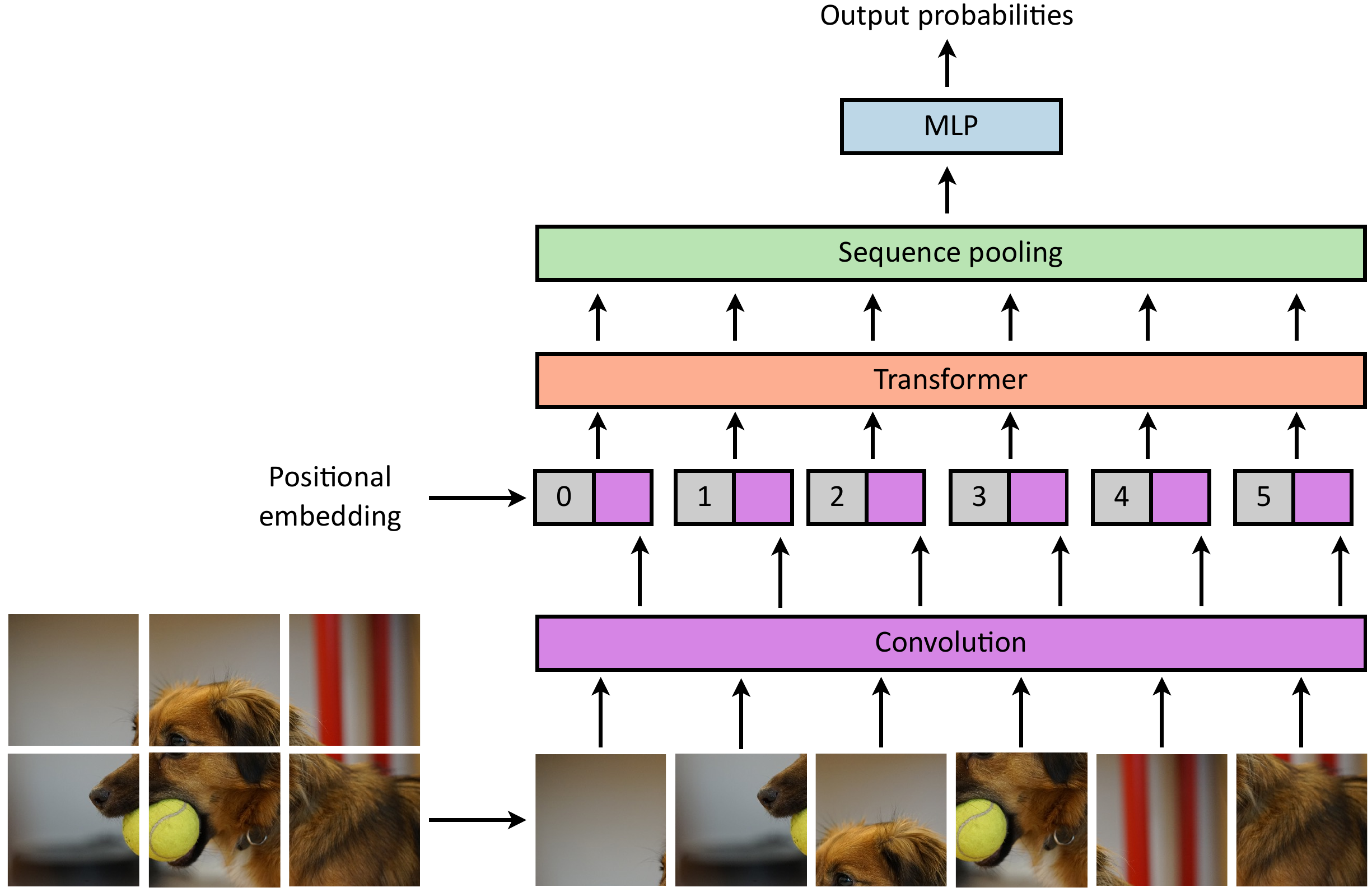}
	   \caption{\textbf{Compact Convolutional Transformers.} This architecture  features a convolutional based patch extraction to leverage a smaller Transformer network, leading to higher data-efficiency. Figure inspired from~\cite{cct}.}
	\label{fig:cct}
\end{figure}

\paragraph{Compact Convolutional Transformer~\cite{cct}.}
DeiT~\cite{deit} and ConVit~\cite{convit} successfully achieve data efficiency at the ImageNet scale. However, ImageNet is a big dataset with 1.3 million images, whereas most datasets are significantly smaller. 

To reach higher data efficiency, the Compact Convolutional Transformer~\cite{cct} uses a CNN operation to extract the patches, and then uses these patches in a Transformer network (Figure~\ref{fig:cct}). 
The Compact Convolutional Transformer comes with some modifications that lead to  major improvements. First, by having a more complex encoding of patches, the system relies on the convolutional inductive biases at the lower scales and then uses a Transformer network to remove the locality constraint of the CNN. Second, the authors show that discarding the `class' token results in higher efficiency. Specifically, instead of the class token, the Compact Convolutional Transformer pools together all the patches token and classifies on top of this pooled token. These two modifications enable using smaller Transformers, while improving both the data efficiency and the computational efficiency. Therefore, these improvements allow the Compact Convolutional Transformer to be successfully trained on smaller datasets, such as CIFAR or MNIST.

\subsection{Computational Efficiency}
\label{sub:compute_efficiency}

The Vision Transformer architecture (Section~\ref{sub:vit}) suffers from a $\mathcal{O}(n^2)$ complexity with respect to the number of tokens. When considering small resolution images or big patch size, this is not a limitation; for instance, for an image of $224 \times 224$ resolution with $16 \times 16$ patches, this amounts to $196$ tokens. However, when needing to process larger images (for instance 3D images in medical imaging) or when considering smaller patches, using and training such models becomes \emph{prohibitive}. 
For instance, in tasks such as segmentation or image generation, it is needed to have more granular representations than $16 \times 16$ patches; hence, it is crucial to solve this issue to enable more applications of Vision Transformer.

\begin{figure}[hbtp]
	\centering
		\includegraphics[width=0.9\textwidth]{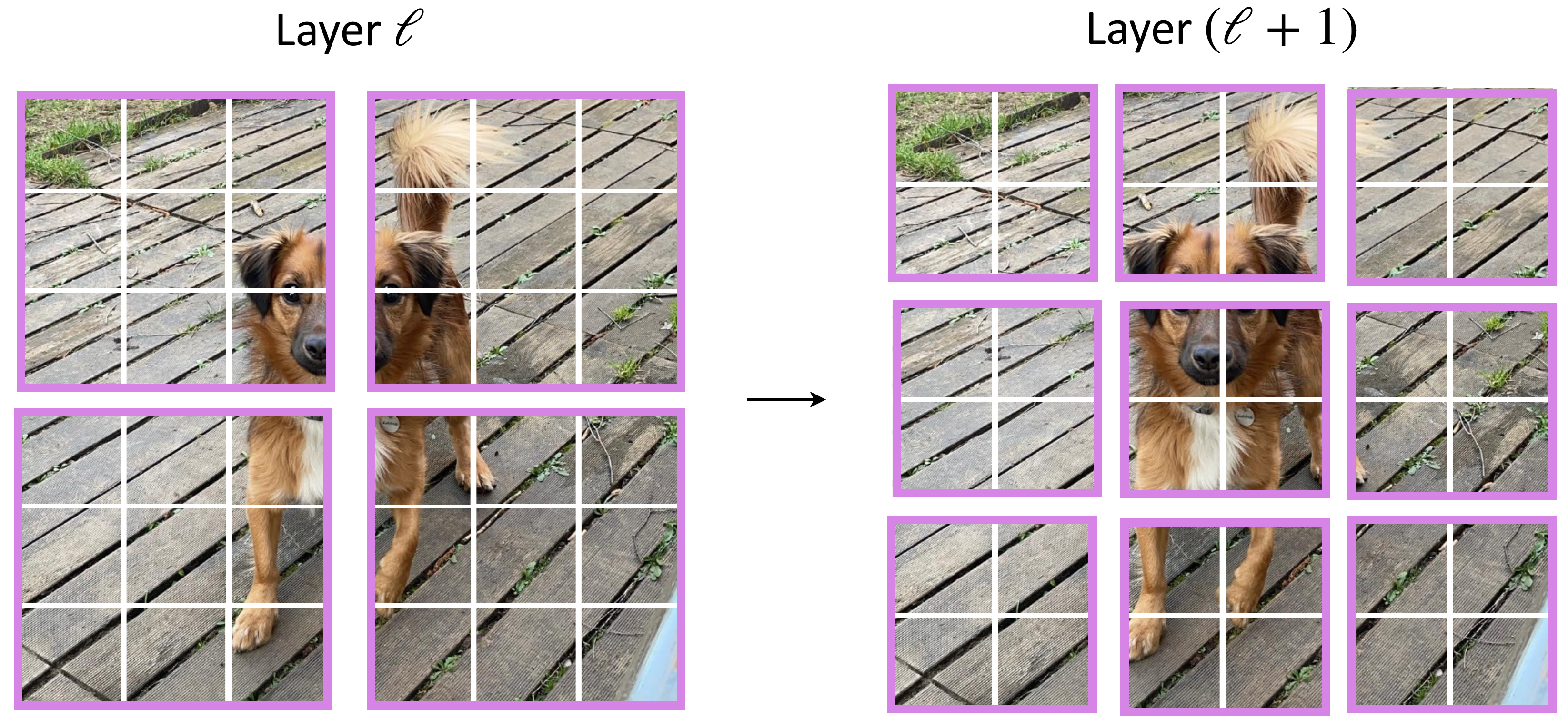}
	   \caption{\textbf{Shifting operation in the Swin Transformer~\cite{swin}.} Between each attention operation, the attention window is shifted so that each patch can communicate with different patches than before. This allows the network to gain more global knowledge with the network's depth. Figure inspired from~\cite{swin}.
	   }
	\label{fig:swin}
\end{figure}

\paragraph{Swin Transformer~\cite{swin}.}
One idea to make Transformers more computation efficient is the Swin Transformer~\cite{swin}. Instead of attending every patch in the image, the Swin Transformer proposes to add a locality constraint. Specifically, the patches can  only attend other patches that are limited to a vicinity window $K$. This restores the local inductive bias of CNNs. To allow communication across patches throughout the network, the Swin Transformer shifts the attention windows from one operation to another (Figure~\ref{fig:swin}). Therefore, the Swin Transformer is quadratic with regard to the size of the window $K$ but linear with respect to the number of tokens $n$ with complexity $\mathcal{O}(nK^2)$. In practice, however, $K$ is small, and this solves the quadratic complexity problem of attention.

\begin{figure}[hbtp]
	\centering
		\includegraphics[width=\textwidth]{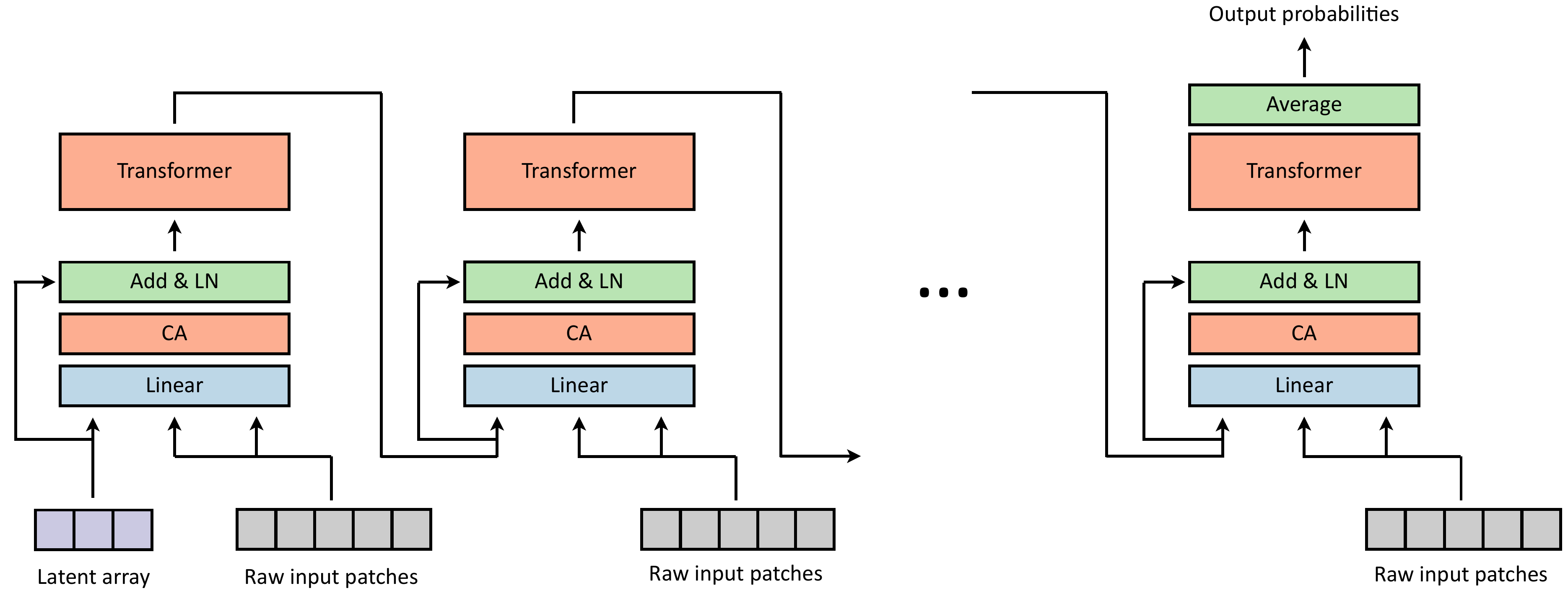}
	   \caption{\textbf{The Perceiver architecture~\cite{jaegle2021perceiver,perceiverio}.} A set of latent tokens retrieve information from the image through Cross-Attention. Self-Attention is performed between the tokens to refine the learned representation. These operations are linear with respect to the number of image tokens. Figure inspired from~\cite{jaegle2021perceiver,perceiverio}.}
	\label{fig:perceiver}
\end{figure}

\paragraph{Perceiver~\cite{jaegle2021perceiver,perceiverio}.}
Another idea for a more computation efficient visual Transformers is to make a more drastic change to the architecture. If instead of using self-attention, the model uses cross-attention, the problem of the quadratic complexity with regard to the number of tokens can be solved. Indeed, computing the cross attention between two sets of length $m$ and $n$, respectively, has complexity $\mathcal{O}(mn)$. This idea is introduced in the Perceiver~\cite{jaegle2021perceiver,perceiverio}. The key idea is to have a smaller set of latent variables that will be used as queries and that will retrieve information in the image token set (Figure~\ref{fig:perceiver}). Since this solves the quadratic complexity issue, it also removes the need of using patches; hence, in the case of Transformers, each pixel is mapped to a single token.

\section{Vision Transformers for tasks other than classification} 
\label{sec:tasks}
Sections~\ref{sec:attention}-\ref{sec:extensions} introduce visual Transformers for one main application: classification. 
Nevertheless, Transformers can be used for numerous other tasks than classification. 

In this section, we present some fundamental vision tasks where Transformers have had a major impact: 
object detection in images (Section~\ref{sub:detection}), image segmentation (Section~\ref{sub:segmentation}), 
training visual Transformers without labels (Section~\ref{sub:no_labels}), and image generation using Generative Adversarial Networks (GANs) (Section~\ref{sub:generative_models}).

\subsection{Object detection with Transformers}
\label{sub:detection}

\begin{figure}[hbtp]
	\centering
		\includegraphics[width=1.0\textwidth]{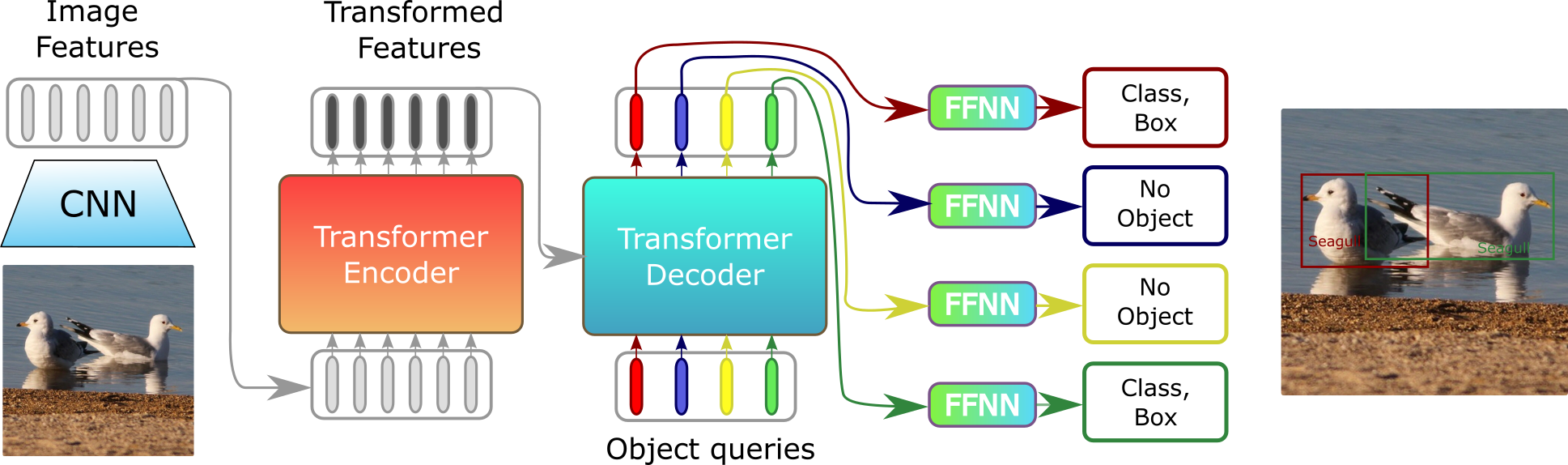}
	   \caption{\textbf{The DETR architecture.} It refines a CNN visual representation to extract object localization and classes. Figure inspired from~\cite{carion2020end}.
	   }
	\label{fig:detr}
\end{figure}

Detection is one of the early tasks that have seen improvements thanks to Transformers. 
Detection is a combined recognition and localization problem; this means that a successful detection system should both recognize whether an object is present in an image and localize it spatially in the image. \cite{carion2020end} is the first approach that uses Transformers for detection. 

\paragraph{DEtection TRansformer (DETR)~\cite{carion2020end}.}
DETR first extracts visual representations with a convolutional network (Figure~\ref{fig:detr})\footnote{Note that in DETR, the Transformer is not directly used to extract the visual representation. Instead, it focuses on refining the visual representation to extract the object information.
}. 
Then, the encodings are processed by a Transformer network. Finally, the processed tokens are provided to a Transformer decoder. The decoder uses cross-attention between a set of learned tokens and the image tokens encoded by the encoder and outputs a set of tokens. Each output token is then passed through a feed-forward network that predicts if an object is present in an image or not; if the object is indeed present, the network also predicts the class and spatial location of the object, i.e., coordinates within the image.

\subsection{Image segmentation with Transformers}
\label{sub:segmentation}

\begin{figure}[hbtp]
	\centering
		\includegraphics[width=0.8\textwidth]{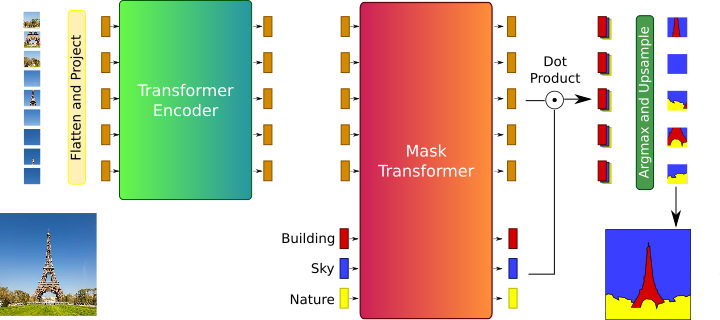} 
	   \caption{\textbf{The Segmenter architecture}. It is a purely ViT based approach to perform semantic segmentation. Figure inspired from~\cite{strudel2021segmenter}.}
	\label{fig:segmenter}
\end{figure}

The goal of image segmentation is to assign to each pixel of an image the label of the object it belongs to. 
The Segmenter~\cite{strudel2021segmenter} is a purely ViT approach addressing image segmentation. The idea is to first use ViT to encode the image. Then, the Segmenter learns a token per semantic label. The encoded patch tokens and the semantic tokens are then fed to a 2nd Transformer. Finally, by computing the scalar product between the semantic tokens and the image tokens, the network assigns a label to each patch. Figure~\ref{fig:segmenter} displays this. 

\subsection{Training Transformers without labels}
\label{sub:no_labels}

Visual Transformers have initially been trained for classification tasks. However, this tasks requires having access to massive amounts of labelled data, which can be hard to obtain (as discussed in Section~\ref{sub:data_efficiency}). Sections~\ref{sub:data_efficiency}-\ref{sub:compute_efficiency} present ways to train ViT more efficiently. However, it would also be interesting to be able to train this type of networks with ``cheaper'' data. Therefore, the goal of this part is to introduce unsupervised learning with Transformers, i.e., training Transformers without any labels. 

\begin{figure}[hbtp]
	\centering
		\includegraphics[width=0.6\textwidth]{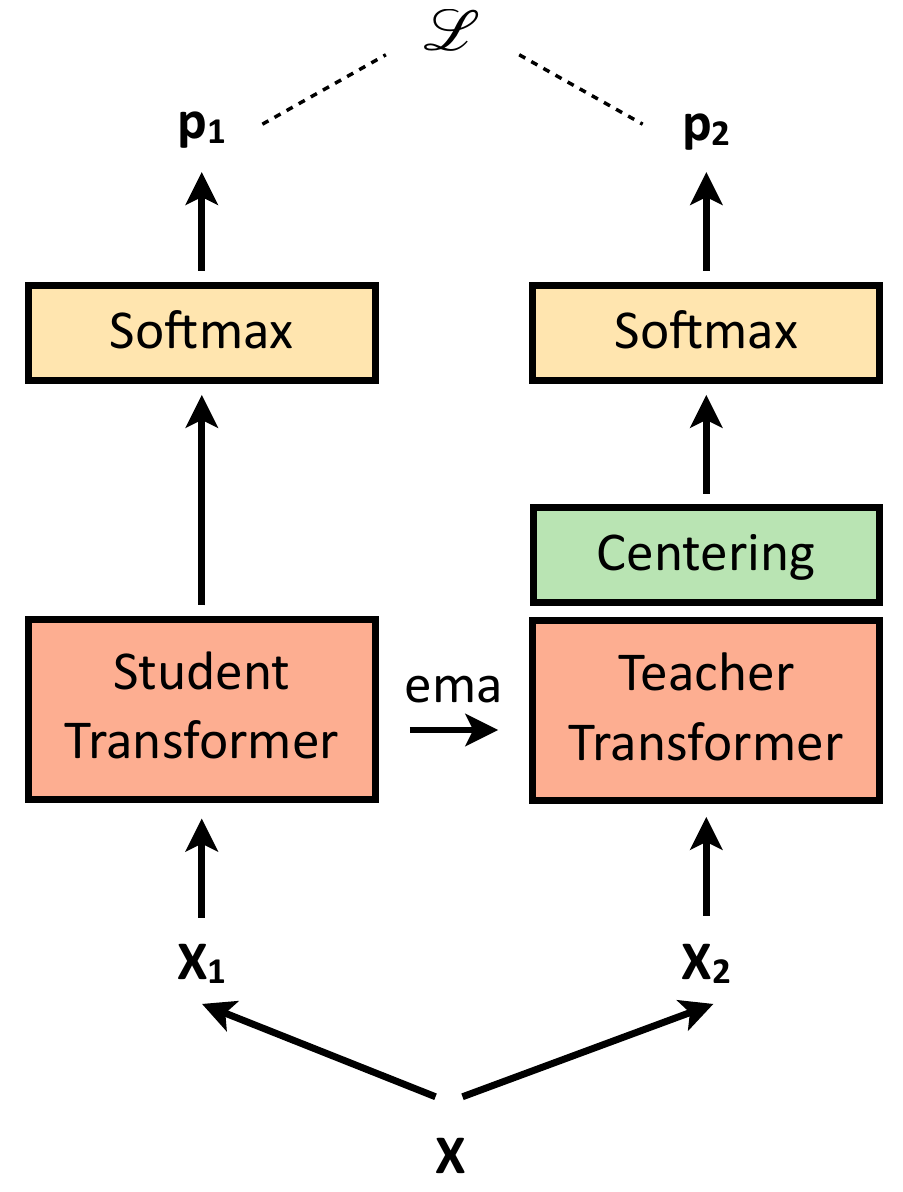}
	   \caption{\textbf{The DINO training procedure.} It consists in matching the outputs between two networks ($\bm{p}_1$ and $\bm{p}_2$) having two different augmentations ($\bm{X}_1$ and $\bm{X}_2$) of the same image as input ($\bm{X}$). The parameters of the teacher model are updated with an exponential moving average (ema) of the student parameters. 
	   Figure inspired from~\cite{dino}.}
	\label{fig:dino_arch}
\end{figure}

\begin{figure}[hbtp]
	\centering
		\includegraphics[width=\textwidth]{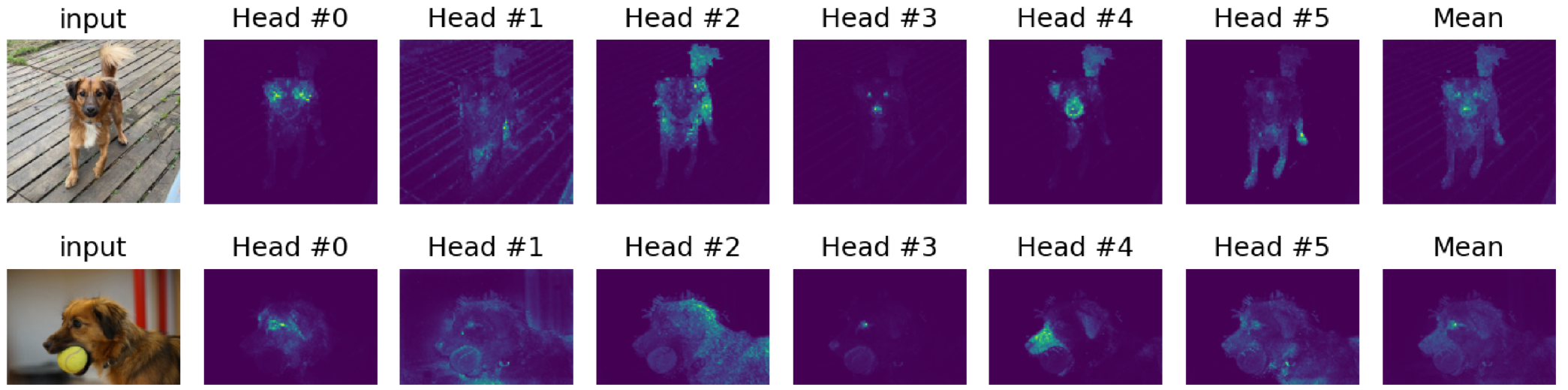}
	   \caption{\textbf{DINO samples.} Visualization of the attention matrix of ViT heads trained with DINO. The ViT discovers the semantic structure of an image in an unsupervised way.}
	\label{fig:dino_sample}
\end{figure}

\paragraph{Self-DIstillation with NO labels (DINO)~\cite{dino}.}
DINO is one of the first works that trains a ViT with self-supervised learning (Figure~\ref{fig:dino_arch}). The main idea is to have two ViT models following the teacher-student paradigm: the first model is updated through gradient descent and the second is an exponential moving average of the first one. 
Then, the whole two-stream DINO network is trained using two augmentations of the same image, that are each passed to one of the two networks. The goal of the training is to match the output between the two networks, i.e., no matter the augmentation in the input data, both networks should produce the same result. 
The main finding of DINO is that the ViT is capable of learning a semantic understanding of the image, as the attention matrices display some semantic information. Figure~\ref{fig:dino_sample} visualizes the attention matrix of the various ViT heads trained with DINO. 

\begin{figure}[hbtp]
	\centering
		\includegraphics[width=\textwidth]{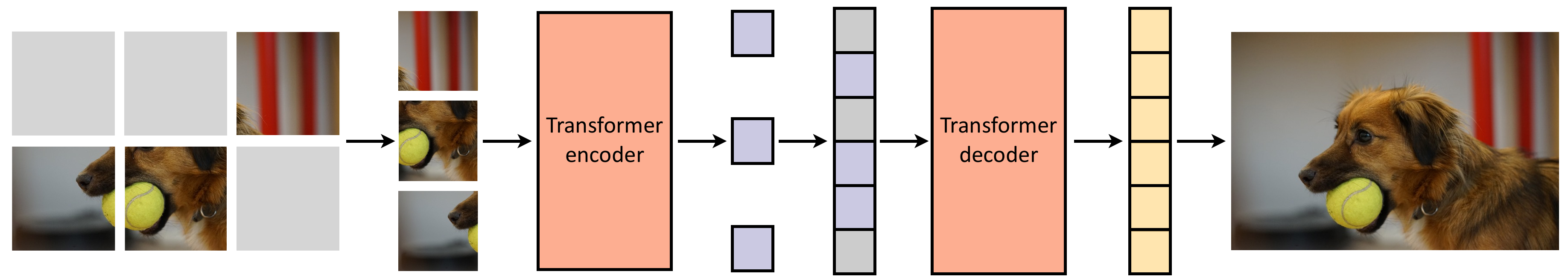}
	   \caption{\textbf{The MAE training procedure}. After masking some tokens of an image, the remaining tokens are fed to an encoder. Then a decoder tries to reconstruct the original image from this representation. 
	   Figure inspired from~\cite{he2021masked}.  }
	\label{fig:mae_arch}
\end{figure}

\paragraph{Masked Autoencoders (MAE)~\cite{he2021masked}.}
Another way to train a ViT without supervision is by using an autoencoder architecture. Masked Autoencoders (MAE)~\cite{he2021masked} perform a random masking of the input token and give the task to reconstruct the original image to a decoder. The encoder learns a representation that performs well in a given downstream task. This is illustrated in Figure~\ref{fig:mae_arch}. 
One of the key observations of the MAE work~\cite{he2021masked} is that the decoder does not need to be very good for the encoder to achieve good performance: by using only a small decoder, MAE successfully trains a ViT in an auto-encoder fashion.

\subsection{Image generation with Transformers and Attention}
\label{sub:generative_models}

Attention and Vision Transformers have also helped in developing fresh ideas and creating new architectures for generative models, and in particular for Generative Adversarial Networks (GANs). 

\begin{figure}[hbtp]
	\centering
		\includegraphics[width=0.4\textwidth]{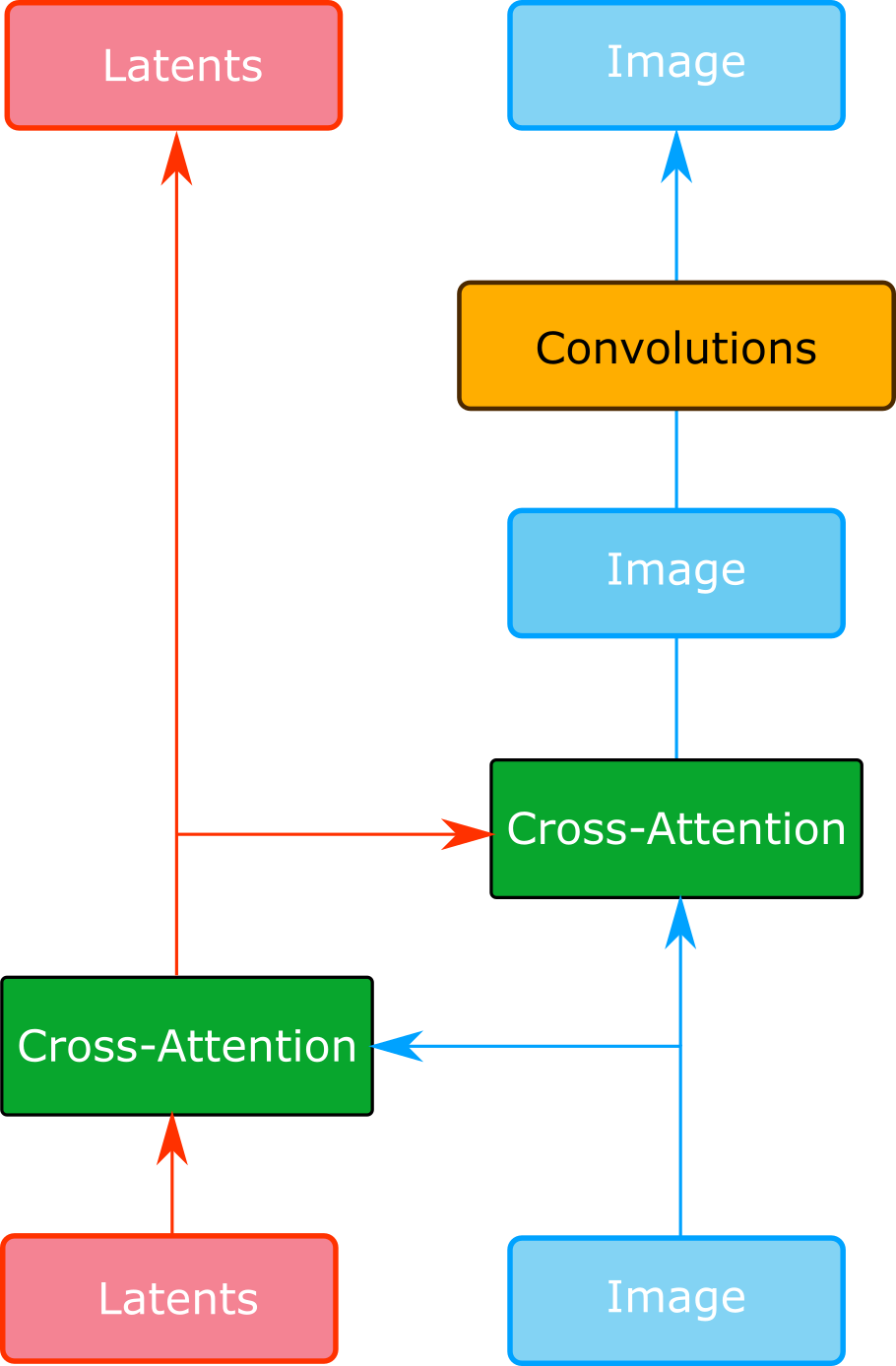}
	   \caption{\textbf{GANsformers architecture}. A set of latents contribute to bring information to a CNN feature map. Figure inspired from~\cite{hudson2021generative}.}
	\label{fig:gansformers}
\end{figure}

\paragraph{GANsformers~\cite{hudson2021generative}.} 
GANsformers are the most representative work of GANs with Transformers, as they are a hybrid architecture using both attention and CNNs. 
The GANsformers architecture is illustrated in Figure~\ref{fig:gansformers}. 
The model first splits the latent vector of a GAN into multiple tokens. Then, a cross attention mechanism is used to improve the generated feature maps and, at the same time, the GANsformers architecture retrieves information from the generated features map to enrich the tokens. This mechanism allows the GAN to have better and richer semantic knowledge, which is showed to be useful for generating multimodal images.

\paragraph{StyleSwin~\cite{zhang2021styleswin}.}
Another approach for generative modeling is to purely use a ViT architecture like StyleSwin~\cite{zhang2021styleswin}. StyleSwin is a GAN that leverages a similar type of attention as the Swin Transformer~\cite{swin}. This allows to generate high definition images without having to deal with the quadratic cost problem.

\section{Vision Transformers for other domains} 
\label{sec:domains}
In this section, we present applications of visual Transformers to other domains. 
First, we describe multimodal Transformers operating with vision and language (Section~\ref{sub:multimodal_transformers}), then we describe video-level attention and video Transformers (Sections~\ref{sub:video_attention}-\ref{sub:video_transformers}), and finally we present multimodal video Transformers operating with vision, language and audio (Section~\ref{sub:multi_modal}). 

\subsection{Multimodal Transformers: vision and language}
\label{sub:multimodal_transformers}

As Transformers have found tremendous success in both natural language processing and computer vision, their use in vision-language tasks is also of interest. In this section, we describe some representative multimodal methods for vision and language: ViLBERT (Section~\ref{subsub:vilbert}), DALL-E (Section~\ref{subsub:dalle}) and CLIP (Section~\ref{subsub:clip}). 

\subsubsection{ViLBERT}
\label{subsub:vilbert}

Vision-and-language BERT (VilBERT)~\cite{lu_vilbert_2019} is an example of architecture that fuses two modalities. 
It consists of two parallel streams, each one working with one modality. The vision stream extracts bounding boxes from images via an object detection network, by encoding their position. The language stream embeds word vectors and extracts feature vectors using the basic Transformers encoder block~\cite{vaswani2017attention} (Figure~\ref{fig:transformer_arch} left). These two resulting feature vectors are then fused together by a Cross-Attention layer (Section~\ref{subsub:cross_attention}). This follows the standard architecture of the Transformers encoder block, where the keys and values of one modality are passed onto the MCA block of the other modality. The output of the Cross-Attention Layer is passed into another Transformers encoder block and these two layers are stacked multiple times. 

The language stream is initialized with BERT trained on Book Corpus~\cite{zhu_corpus_2015} and Wikipedia~\cite{wikidump}, while the visual stream is initialized with Faster R-CNN~\cite{ren_fastercnn(vilbert)_2015}. On top of the pre-training of each stream, the whole architecture is pre-trained on the Conceptual Captions dataset~\cite{sharma2018conceptual} on two pretext tasks. 

ViLBERT has been proven powerful for a variety of multimodal tasks. In the original paper, ViLBERT was fined-tuned to a variety of tasks, including visual question answering, visual common-sense reasoning, referring expressions, and caption-based image retrieval.

\subsubsection{CLIP}
\label{subsub:clip}

Connecting Text and Images (CLIP)~\cite{clip} is designed to address two major issues of deep learning models: costly datasets and inflexibility. While most deep learning models are trained on labelled datasets, CLIP is trained on 400 million text-image pairs that are scraped from the internet. This reduces the labour of having to manually label millions of images that are required to train powerful deep learning models. When models are trained on one specific dataset, they also tend to be difficult to extend to other applications. For instance, the accuracy of a model trained on ImageNet is generally limited to its own dataset and cannot be applied to real-world problems. To optimize training, CLIP models learn to perform a wide variety of tasks during pretraining, and this task allows for zero-shot transfer to many existing datasets. While there are still several potential improvements, this approach is competitive to supervised models that are trained on specific datasets.

\paragraph{CLIP architecture and training.}
CLIP is used to measure the similarity between the text input and the image generated from a latent vector. At the core of the approach is the idea of learning perception from supervision contained in natural language. Methods which work on natural language can learn passively from the supervision contained in the vast amount of text on the internet. 

Given a batch of $N$ (image, text) pairs, CLIP is trained to predict which of the $N \times N$ possible (image, text) pairings across a batch actually occurred. To do this, CLIP learns a multimodal embedding space by jointly training an image encoder and a text encoder to  maximize the cosine similarity of the image and text embeddings of the $N$ real pairs in the batch, while minimizing the cosine similarity of the embeddings of the $N^{2} {\--} N$ incorrect pairings. A symmetric cross entropy loss over these similarity scores is optimized.

Two different architectures were considered for the image encoder. For the first, ResNet-50~\cite{he_resnet50_2017} is used as the base architecture for the image encoder due to its widespread adoption and proven performance. Several modifications were made to the original version of ResNet.
For the second architecture, ViT is used with some minor modifications: first, adding an extra layer normalization to the combined patch and position embeddings before the Transformer, and second, using a slightly different initialization scheme.

The text encoder is a standard Transformer~\cite{vaswani2017attention} (Section~\ref{sub:basic_transformer}) with the architecture modifications described in~\cite{clip}. As a base size, CLIP uses a 63M- parameter 12- layer 512-wide model with 8 attention heads. The Transformer operates on a lower-cased byte pair encoding (BPE) representation of the text with a 49,152 vocab size~\cite{sennrich2015neural}. The max sequence length is capped at 76. The text sequence is bracketed with [SOS] and [EOS] tokens\footnote{[SOS]: start-of-sequence; [EOS]: end-of-sequence} and the activations of the highest layer of the Transformer at the [EOS] token are treated as the feature representation of the text which is layer normalized and then linearly projected into the multimodal embedding space.

\subsubsection{DALL-E and DALL-E 2}
\label{subsub:dalle}

DALL-E~\cite{ramesh_dalle_2021} is another example of the application of Transformers in vision. It generates images from a natural language prompt -- some examples include `an armchair in the shape of an avocado' and `a penguin made of watermelon'. It uses a decoder-only model, which is similar to GPT-3~\cite{brown_gpt3_2020}. DALL-E uses 12 billion parameters and is pre-trained on Conceptual Captions~\cite{sharma2018conceptual} with over 3.3 million text-image pairs. 
DALL-E 2~\cite{ramesh2022dalle2} is the upgraded version of DALL-E, based on diffusion models and CLIP (Section~\ref{subsub:clip}), and allows better performances with more realistic and accurate generated images. 
In addition to producing more realistic results with a better resolution than DALL-E, DALL-E 2 is also able to edit the outputs. Indeed, with DALL-E 2, one can add or remove realistically an element in the output, and can also generate different variations of the same output.
These two models clearly demonstrate the powerful nature and scalability of Transformers that are capable of efficiently processing a web-scale amount of data.

\subsubsection{Flamingo}
\label{subsub:flamingo}

Flamingo~\cite{alayrac2022flamingo} is a visual language model (VLM) tackling a wide range of multimodal tasks based on few-shot learning. This is an adaptation of large language models (LLMs) handling an extra visual modality with 80B parameters.

Flamingo consists of three main components: a vision encoder, a Perceiver resampler and a language model.
First, to encode images or videos, a vision convolutional encoder~\cite{brock2021nfnet} is pretrained in a contrastive way, using image and text pairs~\footnote{The text is encoded using a pretrained BERT model~\cite{devlin2018bert}.}.
Then, inspired by the Perceiver architecture~\cite{jaegle2021perceiver} (detailed in Section~\ref{subsub:cross_attention}), the Perceiver resampler takes a variable number of encoded visual features and outputs a fixed length latent code.
Finally, this visual latent code conditions the language model by querying language tokens through cross attention blocks. Those cross-attention blocks are interleaved with pretrained and frozen language model blocks.

The whole model is trained using three different kinds of datasets without annotations (text with image content from webpages~\cite{alayrac2022flamingo}, text and image pairs~\cite{alayrac2022flamingo, jia2021align} and text and video pairs~\cite{alayrac2022flamingo}).
Once the model is trained, it is fine-tuned using few-shot learning techniques to tackle specific tasks.

\subsection{Video Attention} 
\label{sub:video_attention}

Video understanding is a long-standing problem, and despite incredible Computer Vision advances, obtaining the best video representation is still an active research area. Videos require employing effective spatio-temporal processing of RGB and time streams to capture long-range interactions~\cite{epstein2021learning,marin2019laeo}, while focusing on important video parts~\cite{nagrani2021attention} with minimum computational resources~\cite{ryoo2021tokenlearner}. 

Typically, video understanding benefits from 2D Computer Vision, by adapting 2D image processing methods to 3D spatio-temporal methods~\cite{hara2018cnn3d}.  
And through the Video Vision Transformer (ViViT)~\cite{arnab2021vivit}, history repeats itself. 
Indeed, with the rise of Transformers~\cite{vaswani2017attention} and the recent advances in image classification~\cite{dosovitskiy2020vit}, video Transformers appear as logical successors of CNNs. 

However, in addition to the computationally expensive video processing, Transformers  also  require a lot of computational resources. Thus, developing efficient spatio-temporal attention mechanisms is essential~\cite{arnab2021vivit,bertasius2021timesformer,jaegle2021perceiver}.

In this section, we first describe the general principle of video Transformers (Section~\ref{subsub:general_principle}), and then, we detail three different attention mechanisms used for video representation (Sections~\ref{subsub:full_space_time_attention}, \ref{subsub:divided_space_time_attention} and \ref{subsub:attention_bottleneck}).

\subsubsection{General principle}
\label{subsub:general_principle}

Generally, inputs of video Transformers are RGB video clips $\bm{\mathsfit{X}} \in \mathbb{R}^{F\times H \times W \times 3}$, with $F$ frames of size $H \times W$.

To begin with, video Transformers split the input video clip $\bm{\mathsfit{X}}$ into $ST$ tokens $\bm{x}_i \in \mathbb{R}^K$, where $S$ and $T$ are respectively the number of tokens along the spatial and temporal dimension and $K$ is the size of a token.

To do so, the simplest method extracts non-overlapping 2D patches of size $P\times P$ from each frame~\cite{dosovitskiy2020vit}, as used in TimeSformer~\cite{bertasius2021timesformer}. This results in $S = HW/P^2$, $T=F$ and $K = P^2$.

However, there exist more elegant and efficient token extraction methods for videos. For instance, in ViViT~\cite{arnab2021vivit}, the authors propose to extract 3D volumes from videos (involving $T \neq F$) to capture spatio-temporal information within tokens. In TokenLearner~\cite{ryoo2021tokenlearner}, they propose a learnable token extractor to select the most important parts of the video.

Once raw tokens  $\bm{x}_i$ are extracted, Transformer architectures aim to map them into $d$-dimensional embedding vectors $\bm{Z} \in \mathbb{R}^{ST \times d}$ using a linear embedding $\bm{E} \in \mathbb{R}^{d \times K}$: 
\begin{equation}
    \bm{Z} = [\bm{z}_{cls}, E\bm{x}_1, E\bm{x}_2, \dots, E\bm{x}_{ST}] + \textbf{PE} \quad ,
\end{equation}
where $\bm{z}_{cls} \in \mathbb{R}^{d}$ is a classification token that encodes information from all tokens of a single sample~\cite{devlin2018bert}, and $\textbf{PE} \in \mathbb{R}^{ST \times d}$ is a positional embedding that encodes the spatio-temporal position of tokens, since the subsequent attention blocks are permutation invariant~\cite{vaswani2017attention}.

In the end, embedding vectors $\bm{Z}$ pass through a sequence of $L$ Transformer layers. A Transformer layer $\ell$, is composed of a series of Multi-head Self-Attention (MSA)~\cite{vaswani2017attention}, Layer Normalisation (LN)~\cite{ba2016layernorm} and MLP blocks:
\begin{equation}
\begin{array}{c}
\bm{Y}^\ell = \text{MSA}(\text{LN}(\bm{Z}^\ell)) + \bm{Z}^\ell \quad ,\\
\bm{Z}^{\ell + 1} = \text{MLP}(\text{LN}(\bm{Y}^\ell)) + \bm{Y}^\ell \quad .
\end{array}
\end{equation}

In this way, as shown in Figure~\ref{fig:multi_head_attention}, we denote four different components in a video Transformer layer: the Query-Key-Value (QKV) projection, the MSA block, the MSA projection and the MLP.
For a layer with $h$ heads, the complexity of each component is~\cite{vaswani2017attention}:
\begin{itemize}
    \item \textit{QKV projection}: $\mathcal{O}(h.(2STdd_k + STdd_v)$ 
    \item \textit{MSA}: $\mathcal{O}(hS^2T^2.(d_k + d_v))$ 
    \item \textit{MSA projection}: $\mathcal{O}(SThd_vd)$ 
    \item \textit{MLP}: $\mathcal{O}(STd^2)$ 
\end{itemize}

We note that the MSA complexity is the most impacting component, with a quadratic complexity with respect to the number of tokens. 
Hence, for comprehension and clarity purposes, in the rest of the Section, we consider the global complexity of a video Transformer with $L$ layers to equal to $\mathcal{O}(LS^2T^2)$. 

\begin{figure}[hbtp]
	\centering
	    \includegraphics[width=0.6\textwidth]{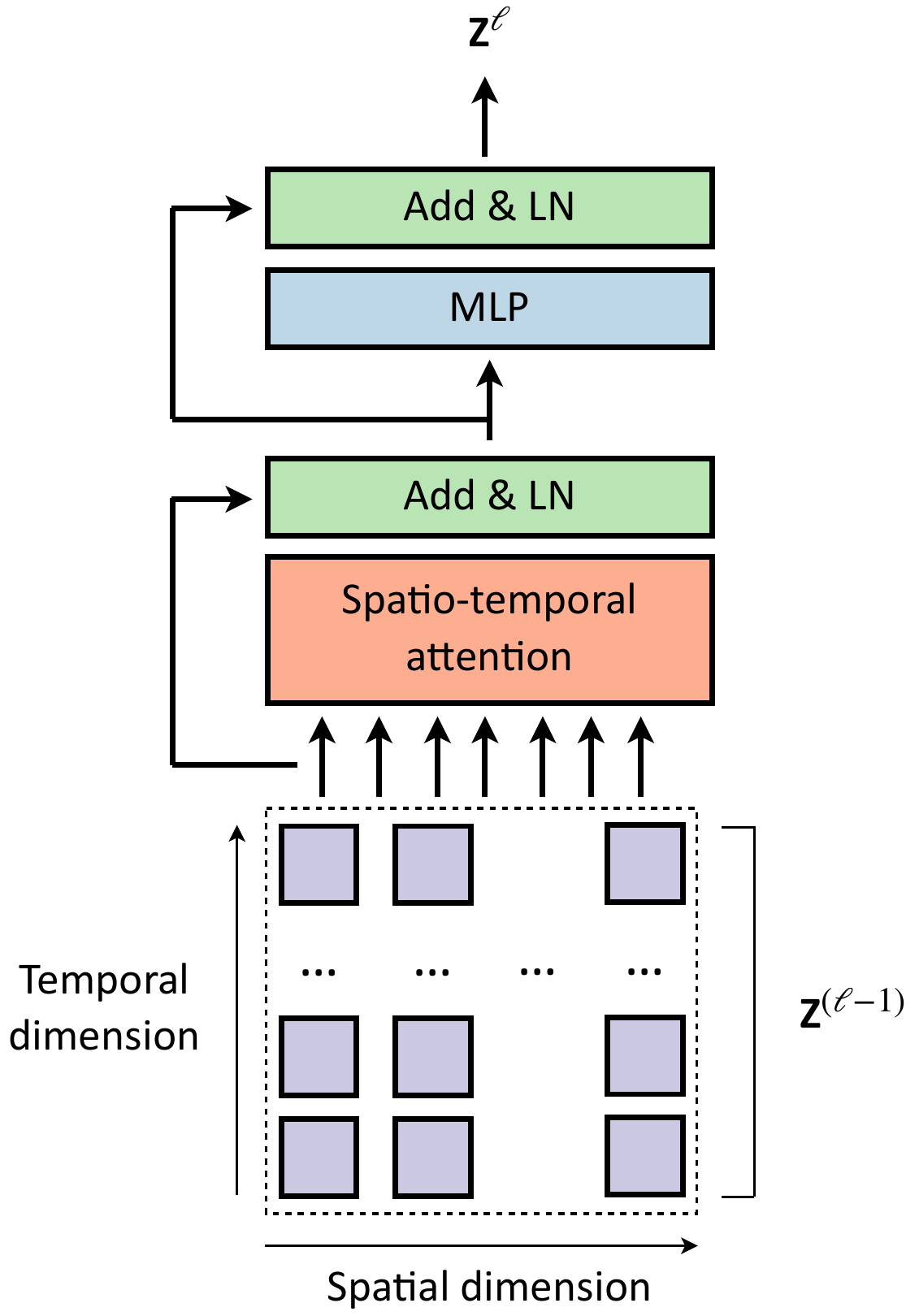}
	    \caption{\textbf{Full space-time attention mechanism.} 
	    Embedding tokens at layer $\ell - 1$, $\bm{Z}^{(\ell -1)}$ are all fed simultaneously through a unique spatio-temporal attention block. Finally, the spatio-temporal embedding is passed through a MLP and normalized to output embedding tokens of the next layer, $\bm{Z}^{\ell}$.
	    Figure inspired from~\cite{bertasius2021timesformer}. }
	\label{fig:full_space_time_attention}
\end{figure}

\subsubsection{Full space-time attention}
\label{subsub:full_space_time_attention}

As described in~\cite{bertasius2021timesformer,arnab2021vivit}, \textit{full space-time attention} mechanism is the most basic and direct spatio-temporal attention mechanism. As shown in Figure~\ref{fig:full_space_time_attention}, it consists in computing self-attention across all pairs of extracted tokens.

This method results in a heavy complexity of $\mathcal{O}(LS^2T^2)$~\cite{bertasius2021timesformer,arnab2021vivit}. This quadratic complexity can fast be memory consuming, which it is especially true when considering videos.
Therefore, using full space-time attention mechanism is  impractical~\cite{bertasius2021timesformer}.  

\begin{figure}[hbtp]
	\centering
	    \includegraphics[width=0.6\textwidth]{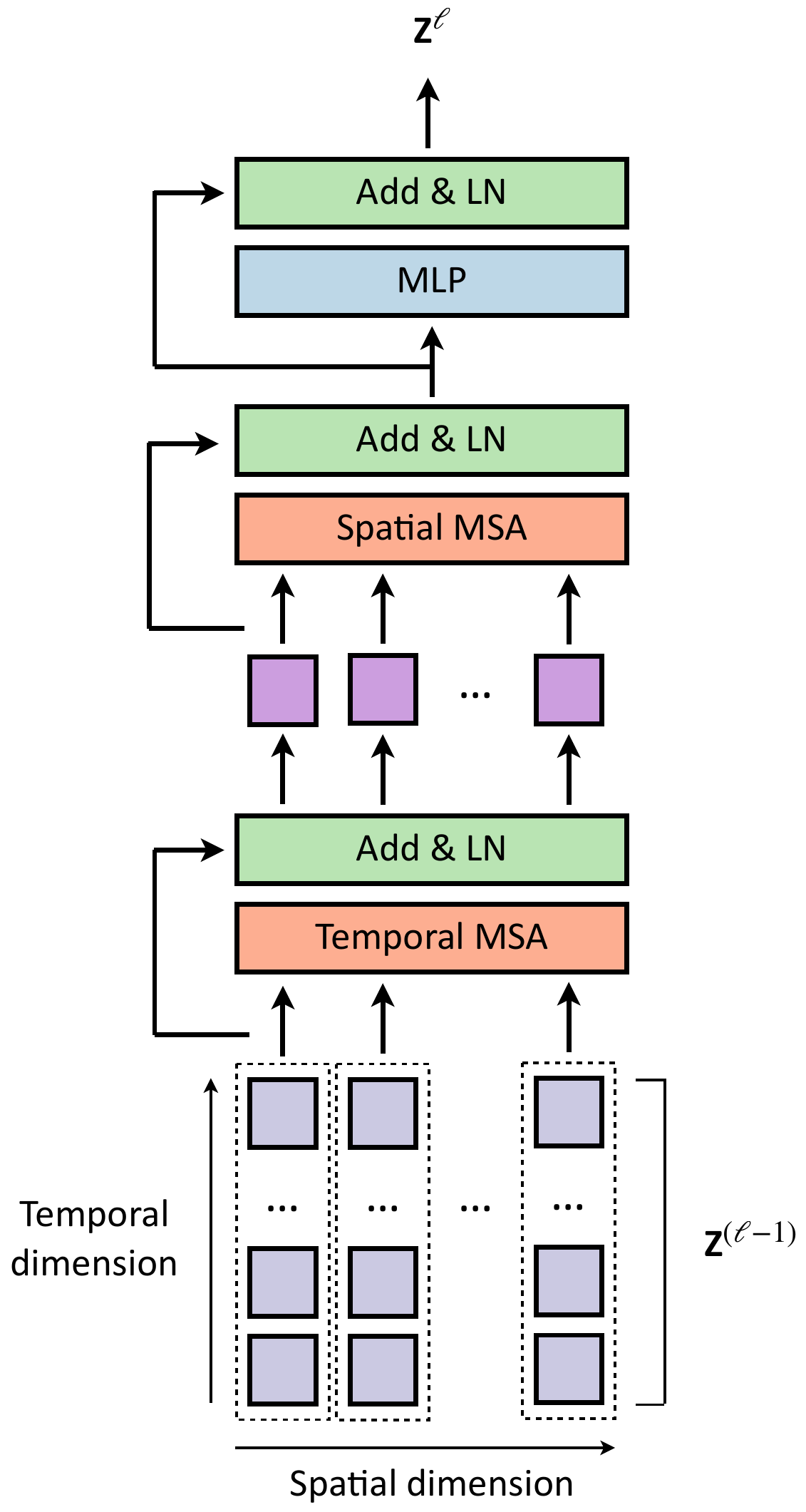}
	    \caption{\textbf{Divided space-time attention mechanism.} 
	    Embedding tokens at layer $\ell - 1$, $\bm{Z}^{(\ell -1)}$ are first processed along the temporal dimension through a first MSA block, and the resulting tokens are processed along the spatial dimension. Finally, the spatio-temporal embedding is passed through a MLP and normalized to output embedding tokens of the next layer, $\bm{Z}^{\ell}$.
	    Figure inspired from~\cite{bertasius2021timesformer}.}
	\label{fig:divided_space_time_attention}
\end{figure}

\subsubsection{Divided space-time attention}
\label{subsub:divided_space_time_attention}

A smarter and more efficient way to compute spatio-temporal attention is the \textit{divided space-time attention} mechanism, first described in~\cite{bertasius2021timesformer}.

As shown in Figure~\ref{fig:divided_space_time_attention}, it relies on computing spatial and temporal attention separately in each Transformer layer. Indeed, we first compute the spatial attention, i.e., self-attention within each temporal index, and then, the temporal attention, i.e., self-attention across all temporal indices.

The complexity of this attention mechanism is $\mathcal{O}(LST.(S+T))$~\cite{bertasius2021timesformer}.
By separating the calculation of the self-attention over the different dimensions, one tames the quadratic complexity of the MSA module. This mechanism highly reduces the complexity of a model with respect to the full space-time complexity. Therefore, it is reasonable to use it to process videos~\cite{bertasius2021timesformer}.

\begin{figure}[hbtp]
	\centering
	    \includegraphics[width=0.6\textwidth]{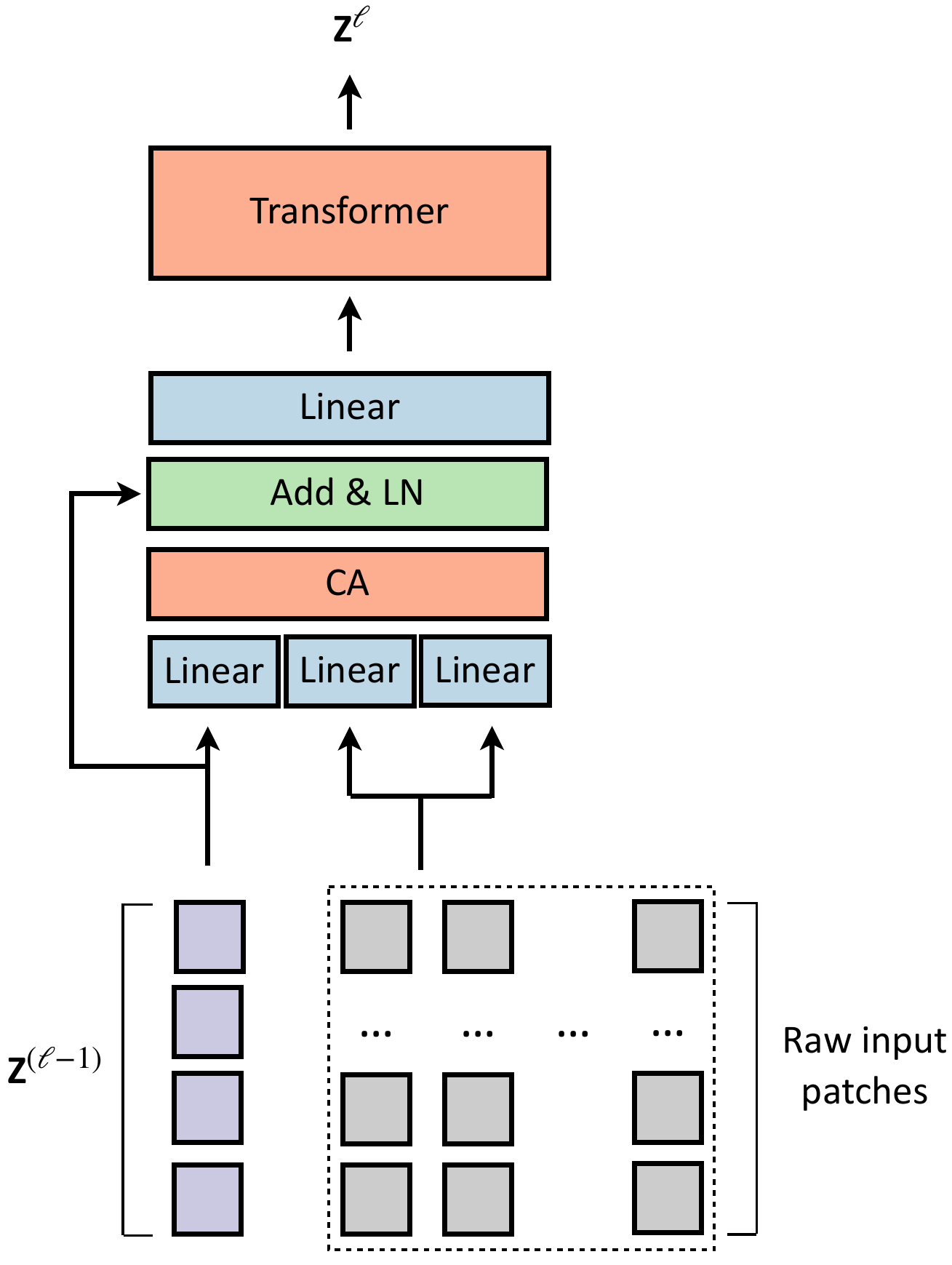}
	    \caption{\textbf{Attention bottleneck mechanism.} 
	    Raw input patches and embedding tokens at layer $\ell - 1$, $\bm{Z}^{(\ell -1)}$ are fed to a cross attention block (CA), then normalized and projected. Finally, the resulting embedding is passed through a transformer to output embedding tokens of the next layer, $\bm{Z}^{\ell}$.
	    Figure inspired from~\cite{jaegle2021perceiver}.}
	\label{fig:attention_bottleneck}
\end{figure}

\subsubsection{Cross-attention bottlenecks}
\label{subsub:attention_bottleneck}

An even more refined way to reduce the computational cost of attention calculation, consists of using cross-attention as a bottleneck. For instance, as shown in Figure~\ref{fig:attention_bottleneck} and mentioned in Section~\ref{sub:compute_efficiency}, the Perceiver~\cite{jaegle2021perceiver} projects the extracted tokens $\bm{x}_i$ into a very low-dimensional embedding through a cross-attention block placed before the Transformer layers. 

Here, the cross attention block placed before the $L$ Transformer layers reduces the input dimension from $ST$ to $N$, where $N \ll ST$~\footnote{In practice, $N \le 512$ for a Perceiver~\cite{jaegle2021perceiver}, against $ST = 16 \times 16 \times (32 / 2) = 4,096$ for a ViViT-L~\cite{arnab2021vivit}}, thus resulting in a complexity of $\mathcal{O}(STN)$. Hence, the total complexity of this attention block is $\mathcal{O}(STN + LN^2)$.
It reduces again the complexity of a model with respect to the \textit{divided space-time attention} mechanism. We note that it enables to design deep architectures, as in the Perceiver~\cite{jaegle2021perceiver}, and then, it enables the extraction of higher level features.

\begin{figure}[hbtp]
	\centering
	    \includegraphics[width=\textwidth]{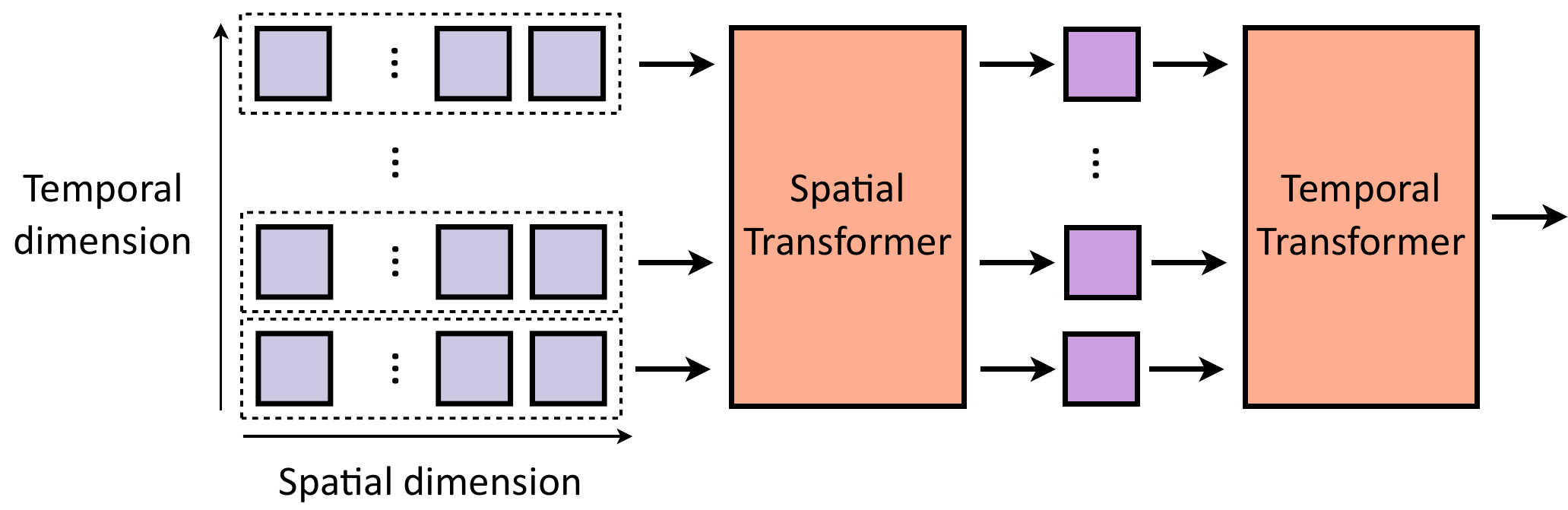}
	    \caption{\textbf{Factorized encoder mechanism.} 
	    First, a spatial Transformer processes input tokens along the spatial dimension, Then, a temporal Transformer processes the resulting spatial embedding along the temporal dimension. 
	    Figure inspired from~\cite{jaegle2021perceiver}.}
	\label{fig:factorised_encoder}
\end{figure}

\subsubsection{Factorised encoder}
\label{subsub:factorised_encoder}

Lastly, the \textit{factorised encoder}~\cite{arnab2021vivit} architecture is the most efficient with respect to the complexity/performance trade-off.

As in \textit{divided space-time attention}, the \textit{factorised encoder} aims to compute spatial and temporal attention separately. Nevertheless, as shown in Figure~\ref{fig:factorised_encoder}, instead of mixing spatio-temporal tokens in each Transformer layer, here there exist two separate encoders. First, a representation of each temporal index is obtained thanks to a spatial encoder with $L_s$ layers, then, these tokens are passed through a temporal encoder with $L_t$ layers (i.e. $L = L_s + L_t$).

Hence, the complexity of a such architecture, has two main components: the spatial encoder complexity of $\mathcal{O}(L_sS^2)$, and the temporal encoder complexity of $\mathcal{O}(L_tT^2)$. It results in a global complexity of $\mathcal{O}(L_sS^2 + L_tT^2)$.
Thus, it leads to very lightweight models. However, as it first extracts per-frame features, and then aggregates them to a final representation, it corresponds to a late fusion mechanism, which can sometimes be a drawback as it does not mix spatial and temporal information simultaneously~\cite{nagrani2021bottlenecksmultimodalfusion}.

\subsection{Video Transformers} 
\label{sub:video_transformers}

In this section, we present two modern Transformer-based architectures for video classification.
We start by introducing the TimeSformer architecture in Section~\ref{subsub:timesformer}, and then the ViViT architecture in Section~\ref{subsub:vivit}.

\subsubsection{TimeSformer}
\label{subsub:timesformer}

\begin{figure}[hbtp]
	\centering
	    \includegraphics[width=0.4\textwidth]{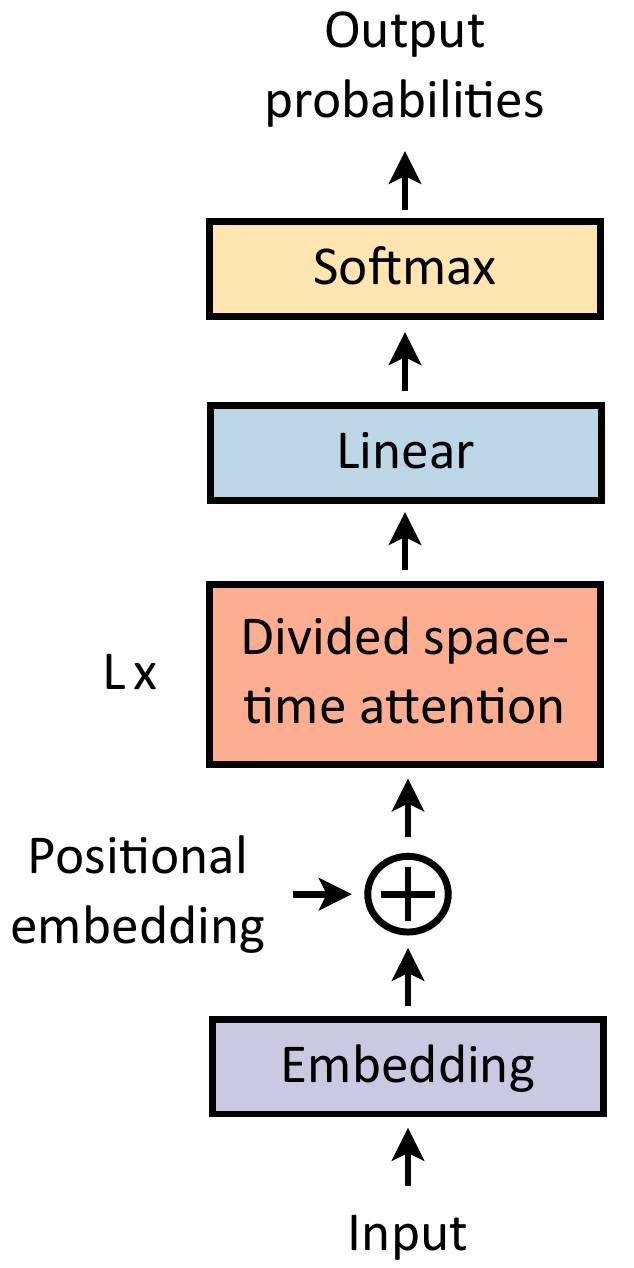}
	    \caption{\textbf{TimeSformer architecture.} 
	    The TimeSformer first projects input to embedding tokens, which are summed to positional embedding tokens. The resulting tokens are then passed through $L$ divided space-time attention blocks and then linearly projected to obtain output probabilities.}
	\label{fig:timesformer}
\end{figure}

TimeSformer~\cite{bertasius2021timesformer} is one of the first architectures with space-time attention that impacted the video classification field. It follows the same structure and principle described in Section~\ref{subsub:general_principle}.

First, it takes as input an RGB video clip sampled at a rate of $1/32$ and decomposed into 2D $16 \times 16$ patches. 

As shown in Figure~\ref{fig:timesformer}, the TimeSformer architecture is based on the ViT architecture (Section~\ref{sub:vit}), with $12$ $12$-headed MSA layers. However, the added value compared to the ViT is that TimeSfomer uses the \textit{divided space-time attention} mechanism (Section~\ref{subsub:divided_space_time_attention}). Such attention mechanism enables to capture high-level spatio-temporal features, while taming the complexity of the model. 
Moreover, the authors introduce three variants of the architecture: (i) TimeSformer, the standard version of the model, that operates on $8$ frames of $224 \times 224$; (ii) TimeSformer-L, a configuration with high spatial resolution, that operates on $16$ frames of $448 \times 448$; and (iii) TimeSformer-HR, a long temporal range setup, that operates on $96$ frames of $224 \times 224$.

Finally, the terminal classification token embedding is passed through an MLP to output a probability for all video classes. During inference, the final prediction is obtained by averaging the output probabilities from three different spatial crops of the input video clip (top-left, centre and bottom-right). 

TimeSformer achieves similar state-of-the-art performances as the 3D CNNs~\cite{feichtenhofer2019slowfast,carreira2017i3d} on various video classification datasets, such as Kinetics-400 and Kinetics-600~\cite{kay2017kinetics}. Note, the TimeSformer is much faster to train (416 training hours against 3840 hours~\cite{bertasius2021timesformer} for a SlowFast architecture~\cite{feichtenhofer2019slowfast}), and also, more efficient ($0.59$ TFLOPs against $1.97$ TFLOPs~\cite{bertasius2021timesformer} for a SlowFast architecture~\cite{feichtenhofer2019slowfast}). 

\subsubsection{ViViT}
\label{subsub:vivit}

\begin{figure}[hbtp]
	\centering
	    \includegraphics[width=0.4\textwidth]{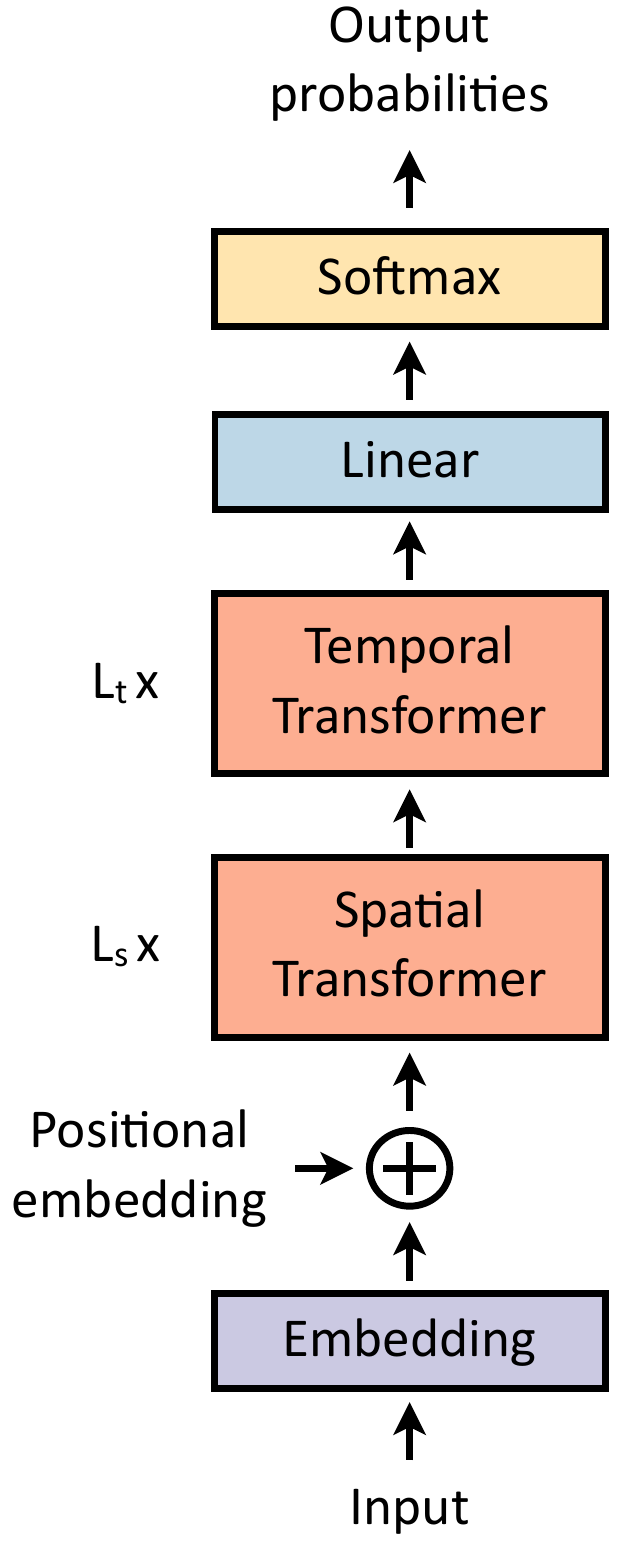}
	    \caption{\textbf{ViViT architecture.}
	    The ViViT first projects input to embedding tokens, which are summed to positional embedding tokens. The resulting tokens are first passed through $L_s$ spatial attention blocks and then through $L_t$ temporal attention blocks. The resulting output is linearly projected to obtain output probabilities.
	    }
	\label{fig:vivit}
\end{figure}

ViViT~\cite{arnab2021vivit} is the main extension of the ViT~\cite{dosovitskiy2020vit} architecture (Section~\ref{sub:vit})  for video classification.

First, the authors use a $16 \time 16 \time 2$ tubelet embedding instead of a 2D patch embedding, as mentioned in Section~\ref{subsub:general_principle}. This alternate embedding method aims to capture the spatio-temporal information from the tokenization step, unlike standard architectures that fuse spatio-temporal information from the first attention block.

As shown in Figure~\ref{fig:vivit}, the ViViT architecture is based on \textit{factorised encoder} architecture (Section~\ref{subsub:factorised_encoder}), and consists of one spatial and one temporal encoder operating on input clips with $32$ frames of $224 \times 224$. 
The spatial encoder uses one of the three ViT variants as backbone~\footnote{ViT-B: $12$ $12$-headed MSA layers; ViT-L: $24$ $16$-headed MSA layers; and ViT-H: $32$ $16$-headed MSA layers.}. For the temporal encoder, the number of layers does not impact much the performance, so that, according to the performance/complexity trade-off, the number MSA layers is fixed at $4$. 
The authors show that such architecture reaches high performances, while reducing drastically the complexity.

Finally, as in TimeSformer (Section~\ref{subsub:timesformer}), ViViT outputs probabilities for all video classes through the last classification token embedding, and averages the obtained probabilities across three crops of each input clip (top-left, centre and bottom-right).

ViViT outperforms both 3D CNNs~\cite{feichtenhofer2019slowfast,carreira2017i3d} and TimeSformer~\cite{bertasius2021timesformer} on the Kinetics-400 and Kinetics-600 datasets~\cite{kay2017kinetics}. 
Note, the complexity of this architecture is highly reduced in comparison to other state-of-the-art models. For instance, the number of FLOPs for a ViViT-L/$16 \times 16 \times 2$ is $3.89 \times 10^{12}$, against $7.14 \times 10^{12}$ for a TimeSformer-L~\cite{bertasius2021timesformer}, and $7.14 \times 10^{12}$ for a SlowFast~\cite{feichtenhofer2019slowfast} architecture.

\subsection{Multimodal video Transformers} 
\label{sub:multi_modal}

Nowadays, one of the main gaps between artificial and human intelligence is the ability for us to process multimodal signals and to enrich the analysis by mixing the different modalities. Moreover, until recently, deep learning models have been focusing mostly on very specific visual tasks, typically based on a single modality, such as image classification~\cite{deng2009imagenet,krizhevsky2009cifar,dosovitskiy2020vit,wu2021cvt,touvron2021cait}, audio classification~\cite{gemmeke2017audioset,gong2021ast,nagrani2021bottlenecksmultimodalfusion,jaegle2021perceiver} and machine translation~\cite{bojar2014wmt,devlin2018bert,liu2020transformerBT,edunov2018backtranslation,lin2020pre}. These two factors combined have pushed researchers to take up multimodal challenges.

The \textit{default} solution for multimodal tasks consists in first creating an individual model (or network) per modality, and then in fusing the resulting single-modal features together~\cite{owens2018audio,ramanishka2016multimodal}.  
Yet, this approach fails to model interactions or correlations among different modalities.
However, the recent rise of attention~\cite{vaswani2017attention,dosovitskiy2020vit,arnab2021vivit} is promising for multimodal applications, since attention performs very well at combining multiple inputs~\cite{chen2022mmvit,jaegle2021perceiver,nagrani2021bottlenecksmultimodalfusion,narasimhan2021clipit}.
\\

Here, we present two main ways of dealing with several modalities: 

\paragraph{1. Concatenating tokens from different modalities into one vector~\cite{chen2022mmvit,jaegle2021perceiver}.}
The multimodal Video Transformer (MM-ViT)~\cite{chen2022mmvit} combines raw RGB frames, motion features and audio spectrogram for video action recognition. To do so, the authors fuse tokens from all different modalities into a single input embedding and pass it through Transformer layers. 
However, a drawback of this method is that it fails to distinguish well one modality to another. 
To overcome this issue, the authors of the Perceiver~\cite{jaegle2021perceiver} propose to learn a modality embedding in addition to the positional embedding (see Sections~\ref{sub:compute_efficiency} and \ref{subsub:general_principle}). This allows associating each token with its modality. 
Nevertheless, given that (i) the complexity of a Transformer layer is quadratic with respect to the number of tokens  (Section~\ref{subsub:general_principle}), and (ii) with this method, the number of tokens is multiplied by the number of modalities: it may lead to skyrocketing computational cost~\cite{chen2022mmvit}.

\paragraph{2. Exploiting cross attention~\cite{nagrani2021bottlenecksmultimodalfusion,narasimhan2021clipit,zadeh2019factorized}.}
Several modern approaches exploit cross attention to mix multiple modalities, such as~\cite{nagrani2021bottlenecksmultimodalfusion} for audio and video, \cite{narasimhan2021clipit} for text and video, and \cite{zadeh2019factorized} for audio, text and video. 
The commonality among all these methods is that they exploit the intrinsic properties of cross attention by querying one modality with a key-value pair from the other one~\cite{nagrani2021bottlenecksmultimodalfusion,narasimhan2021clipit}. 
This idea can be easily generalized to more than two modalities by computing cross-attention across each combination of modalities~\cite{zadeh2019factorized}.

\section{Conclusion} 
\label{sec:conclusion}
Attention is an intuitive and efficient technique that enables handling local and global cues.

On this basis, the first pure attention architecture, the Transformer~\cite{vaswani2017attention}, has been designed for NLP purposes. Quickly, the Computer Vision field has adapted the Transformer architecture for image classification, by designing the first visual Transformer model: the Vision Transformer (ViT)~\cite{dosovitskiy2020vit}.

However, even if Transformers naturally lead  to high performances, the raw attention mechanism is a computationally greedy and heavy technique. For this reason, several enhanced and refined derivatives of attention mechanisms have been proposed~\cite{deit,convit,cct,swin,perceiverio,jaegle2021perceiver}.

Then, rapidly, a wide variety of other tasks have been conquered by Transformer-based architectures, such as object detection~\cite{carion2020end}, image segmentation~\cite{strudel2021segmenter}, self-supervised learning~\cite{dino,he2021masked} and image generation~\cite{hudson2021generative,zhang2021styleswin}.
In addition, Transformer-based architectures are particularly well suited to handle multidimensional tasks.
This is because multimodal signals are easily combined through attention blocks, in particular vision and language cues~\cite{lu_vilbert_2019,ramesh_dalle_2021,clip} and spatio-temporal signals are also easily tamed, as in~\cite{bertasius2021timesformer,arnab2021vivit,jaegle2021perceiver}.

For these reasons, Transformer-based architectures enabled many fields to make tremendous progresses in the last few years.
In the future, Transformers will need to become more and more computationally efficient, e.g. to be usable on cellphones, and will play a huge role to tackle multimodal challenges and bridge together most AI fields. 

\bibliographystyle{spbasicsort}
\bibliography{longstrings,main}

\end{document}